\documentclass{article}

\PassOptionsToPackage{numbers, compress}{natbib}
\usepackage[eandd, final]{neurips_2026}

\usepackage[utf8]{inputenc} 
\usepackage[T1]{fontenc}    
\usepackage{hyperref}       
\usepackage{url}            
\usepackage{booktabs}       
\usepackage{amsmath}        
\usepackage{amsfonts}       
\usepackage{nicefrac}       
\usepackage{microtype}      
\usepackage[table]{xcolor}  
\usepackage{todonotes}      
\usepackage{subcaption}
\usepackage{tabularx}
\usepackage{graphicx}
\usepackage{xparse}
\usepackage{multirow}
\usepackage{wrapfig}
\usepackage{makecell}

\definecolor{tmBlue}{HTML}{1F4E79}
\definecolor{tmTeal}{HTML}{1B7F79}
\definecolor{tmGray}{HTML}{F4F6F8}
\definecolor{tmGreen}{HTML}{2E7D32}
\definecolor{tmOrange}{HTML}{C77700}

\newcolumntype{C}{>{\centering\arraybackslash}X}
\newcolumntype{R}{>{\raggedleft\arraybackslash}X}
\newcommand{\rotate}[1]{\makebox[1em][c]{\rotatebox{90}{#1}}\hfill\null}

\title{TerraMesh-Masks: Open‑Vocabulary\\Segmentation for Earth Observation}

\author{%
  Benedikt Blumenstiel\thanks{Equal contributions} \\
  IBM Research \\
  ETH Zurich \\
  Zurich, Switzerland \\
  \texttt{benedikt.blumenstiel@ibm.com} \\
  \And
  Hugues Devimeux$^*$ \\
  IBM Research \\
  ETH Zurich \\
  Zurich, Switzerland \\
  \texttt{hdevimeux@student.ethz.ch} \\
  \And
  Johannes Jakubik \\
  IBM Research \\
  Zurich, Switzerland \\
  \texttt{johannes.jakubik1@ibm.com} \\
  \And
  Konrad Schindler \\
  ETH Zurich \\
  Zurich, Switzerland \\
  \texttt{schindler@ethz.ch} \\
} 

\begin{document}


\maketitle


\begin{abstract}
Open‑vocabulary segmentation (OVS) enables models to segment a variety of concepts specified in natural language, removing the need for fixed label sets and fine-tuning. While recent OVS approaches have shown strong performance on natural images, they transfer poorly to Earth observation (EO) data due to differing semantics, scales, sensing modality, and geographic context. To date, large‑scale EO‑specific datasets and benchmarks for OVS remain scarce.
We, therefore, introduce \textit{TerraMesh-Masks}, a large-scale open-vocabulary segmentation dataset for EO, comprising 16 million binary masks aligned with nine million multimodal TerraMesh samples and annotated using OpenStreetMap and Overture-derived semantics. In addition, we provide an expert-reviewed evaluation set of 523 samples covering over 200 classes to enable robust evaluation.
To demonstrate the impact of TerraMesh-Masks, we build an open-vocabulary version of TerraMind. Our results demonstrate significantly improved performance over general-purpose state-of-the-art models reaching 35.18\% IoU versus 16.29\% for DINOv3.txt and 11.33\% for SAM-3. We release the dataset publicly under a permissive license at \url{https://huggingface.co/datasets/ibm-esa-geospatial/TerraMesh-Masks}.
\end{abstract}

\section{Introduction}

Open-vocabulary segmentation (OVS) promises more flexible and scalable interaction with Earth observation (EO) data compared to fine-tuning models for a fixed set of classes manually. Users could query models with flexible language rather than a pre-defined closed label set, and use a single model for a variety of concepts~\cite{carion2025sam}. However, existing OVS models are developed primarily for natural images and transfer poorly to satellite imagery~\cite{blumenstiel2023mess}, where semantic concepts are expressed at lower spatial resolution, often require multispectral sensing, and vary strongly across geography and land-cover context. While previous works focused on high-resolution aerial imagery~\cite{shabbir2025geopixel,yao2025remotesam,zheng2025instructsam}, the field currently lacks both a strong benchmark and a large-scale training resource for open-vocabulary segmentation with satellite data.

\begin{figure}
    \centering
    \vspace{-3mm}
    \includegraphics[width=0.7\linewidth]{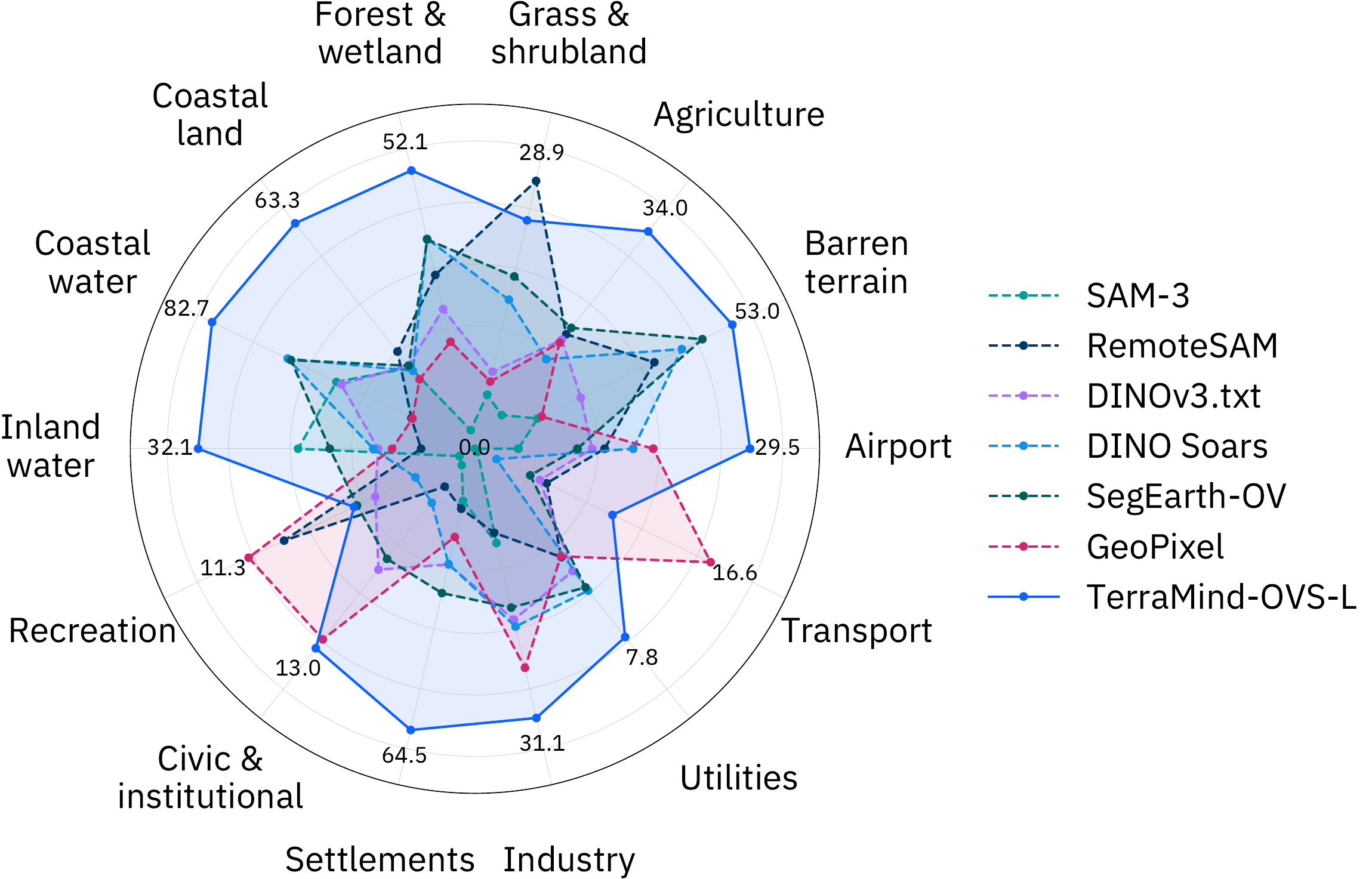}
    \caption{Average IoU$_{pos}$ results per group on TerraMesh-Masks-Eval. While other models struggle with most categories, TerraMind-OVS performs particularly well on tasks related to land cover. Building-related categories are generally challenging due to many small masks which are best identified by GeoPixel.}
    \label{fig:model_results}
    \vspace{-3mm}
\end{figure}

We address this gap with TerraMesh-Masks, a planetary-scale semantic-mask dataset, grounded in OpenStreetMap and Overture Maps, built on top of TerraMesh~\cite{blumenstiel2025terramesh}. We additionally provide an expert-reviewed evaluation set called TerraMesh-Masks-Eval for benchmarking future OVS models. To demonstrate the value of this dataset, we adapt TerraMind to open-vocabulary binary segmentation. Our results show that EO-specific multimodal training is substantially more effective than applying a general-purpose baseline off the shelf. As shown in Fig.~\ref{fig:model_results}, TerraMind-OVS models outperform the generic models SAM-3~\cite{carion2025sam} and DINOv3.txt~\cite{simeoni2025dinov3} as well as many EO-specific models like GeoPixel~\cite{shabbir2025geopixel} in the vast majority of categories.

We make the following contributions:
(1)~We introduce TerraMesh-Masks, a large-scale open-vocabulary segmentation dataset for EO comprising 16 million multimodal image-mask samples with broad geographic and semantic diversity;
(2)~We provide an expert-reviewed evaluation set, enabling grounded benchmarking of OVS models on satellite imagery; and
(3)~We extend TerraMind to open-vocabulary segmentation as an empirical demonstration of the impact of TerraMesh-Masks, revealing substantially better performance compared to the state-of-the-art.

\section{Related work}

Vision--language models (VLMs), such as CLIP~\cite{radford2021clip}, establish robust joint image--text representations that represent the foundation of many open-vocabulary segmentation approaches. Building on these representations, early systems such as CAT-Seg~\cite{cho2024catseg} and SAN~\cite{xu2023san} demonstrated the feasibility of OVS for natural images. More recently, models including SAM-3~\cite{carion2025sam} and DINOv3.txt~\cite{simeoni2025dinov3} have been explicitly designed for spatial segmentation in open-vocabulary settings. Despite their broad training, generalist models often struggle to transfer to domain-specific inputs~\cite{blumenstiel2023mess}.

Compared to natural images, remote sensing imagery exhibits strong variability in spatial resolution, viewpoint, seasonal effects, and geographic distribution. Moreover, many semantic classes are only weakly expressed in RGB imagery. To address these issues, several works adapt VLMs to the EO domain.
MS-CLIP~\cite{marimo2025beyond} leverages LLM-generated captions for multispectral imagery, demonstrating significant performance gains when incorporating additional spectral bands. SegEarth-OV~\cite{li2025segearthov}, DINO Soars~\cite{faulkenberry2026dinosoars}, and other approaches~\cite{sosa2026enabling,zheng2025instructsam} focus on training-free adaptation of CLIP, SAM, or DINO to enable OVS in EO, with strong results on high-resolution imagery.
Beyond training-free methods, OVRSISS~\cite{ye2025ovrs} and AerOSeg~\cite{dutta2025aeroseg} introduce high-resolution datasets to fine-tune CLIP- and SAM-based models for aerial imagery. Similarly, MM-OVSeg~\cite{wei2026mmovseg} employs 25k multimodal samples for supervised adaptation. Multimodal LLM-based systems such as GeoPixel~\cite{shabbir2025geopixel} and RemoteSAM~\cite{yao2025remotesam} extend this paradigm to broader vision--language tasks. EarthMind~\cite{shu2025earthmind} further scales training by combining six public segmentation datasets with labels generated by GPT-4o.

\begin{table}[h]
    \centering
    \setlength\tabcolsep{3pt}
    \footnotesize
    \caption{Overview of existing remote sensing datasets for OVS. Most datasets focus on high-resolution imagery. Some dataset licenses might not reflect source dataset restrictions.}
    \label{tab:datasets}
    \begin{tabularx}{\textwidth}{l C C c c c c c}
    \toprule
    Dataset & Number of images & Vocabulary size & Modalities & Resolution & Image source & \makecell{Annotation\\source} & License \\
    \midrule
    GeoPixelD~\cite{shabbir2025geopixel} & 18.7k & 53k & RGB & $\sim$1m & iSAID & Source labels & CC-BY 4.0 \\
    MM-OVSeg~\cite{wei2026mmovseg} & 25k & unknown & RGB, SAR & 0.5--3 m & \makecell{SpaceNet6;\\DFC2023} & unknown & unknown \\
    LandDiscover50K~\cite{ye2025ovrs} & 51k & 40 & RGB & high-res & 9 datasets & Source labels & MIT \\ 
    RemoteSAM-270K~\cite{yao2025remotesam} & 270k & 1000 & RGB & high-res & 10 datasets & Source labels & CC-BY 4.0 \\
    EarthMind~\cite{shu2025earthmind} & >500k & unknown & Optical, SAR & various & 7+ datasets & Source labels & Apache 2.0 \\\midrule
    TerraMesh-Masks (ours) & 9M & 44k & \makecell{S1, S2,\\RGB, DEM} & 10m & TerraMesh & \makecell{Overture,\\OpenStreetMap} & ODbL \\
    \bottomrule
     \end{tabularx}
\end{table}

Tab.~\ref{tab:datasets} summarizes existing EO datasets for OVS. Other datasets often focus on high-resolution imagery, cover a limited set of semantic classes, or are proprietary and not publicly available~\cite{ye2025ovrs,shabbir2025geopixel,yao2025remotesam,shu2025earthmind,wei2026mmovseg,googledataset}. Additionally, existing models are often evaluated on benchmarks dominated by a small number of land cover categories or high-resolution object classes.

TerraMesh-Masks is designed to address these limitations by providing a large-scale multimodal training dataset and a semantically diverse evaluation benchmark. To the best of our knowledge, TerraMesh-Masks is the first dataset that extends a large-scale unlabeled pretraining dataset to open-vocabulary segmentation by grounding images in OpenStreetMap and Overture data and adding annotations from a large language model (LLM), enabling substantially broader semantic coverage and an order of magnitude more training samples compared to prior work.

\section{TerraMesh-Masks}
\label{sec:dataset}


\begin{figure}[b]
  \centering
   \includegraphics[width=\textwidth]{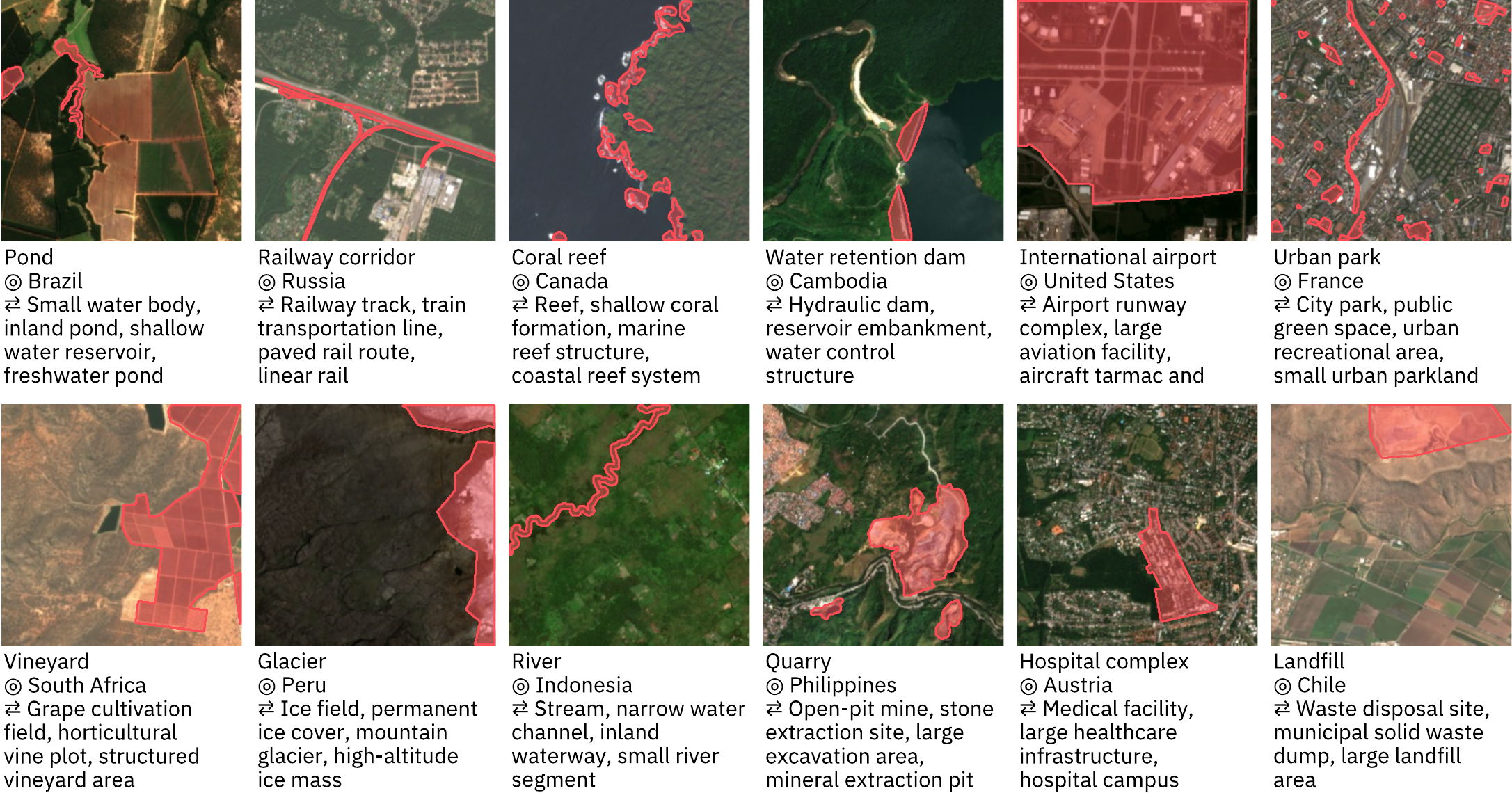}
  \caption{TerraMesh-Masks samples drawn from the manually reviewed Eval set. Each patch is shown with a mask overlay and its open-vocabulary class label, country, and a synonym set.}
  \label{fig:dataset-overview}
\end{figure}

TerraMesh-Masks forms a planetary-scale, multimodal, analysis-ready dataset for open-vocabulary segmentation in Earth observation, built on top of TerraMesh~\cite{blumenstiel2025terramesh}. TerraMesh itself combines data from SSL4EO-S12\,v1.1~\cite{blumenstiel2025ssl4eo} and MajorTOM~\cite{francis2024major}. The dataset provides over nine million globally distributed co-registered patches at 10\,m resolution across modalities such as Sentinel-1 SAR, Sentinel-2 optical, DEM, NDVI, and land-cover layers~\cite{blumenstiel2025terramesh}. TerraMesh-Masks complements the original TerraMesh dataset by adding binary semantic masks with language annotations. In total, TerraMesh-Masks contains roughly 16 million masks derived from OpenStreetMap and Overture Maps, making it, to the best of our knowledge, one of the largest resources for segmentation in Earth observation. More importantly, TerraMesh-Masks has been explicitly designed for open-vocabulary learning. TerraMesh-Masks inherits the broad geographic diversity of TerraMesh and covers 174 countries spanning 22 continental regions. Rather than concentrating on a small set of curated benchmark geographies, it exposes models to diverse landscapes, climates, and built environments at planetary scale. Examples from the dataset are shown in Fig.~\ref{fig:dataset-overview}.

\paragraph{Dataset creation overview.}
\label{sup:dataset_creation}

\begin{figure}[h]
    \centering
    \includegraphics[width=\linewidth]{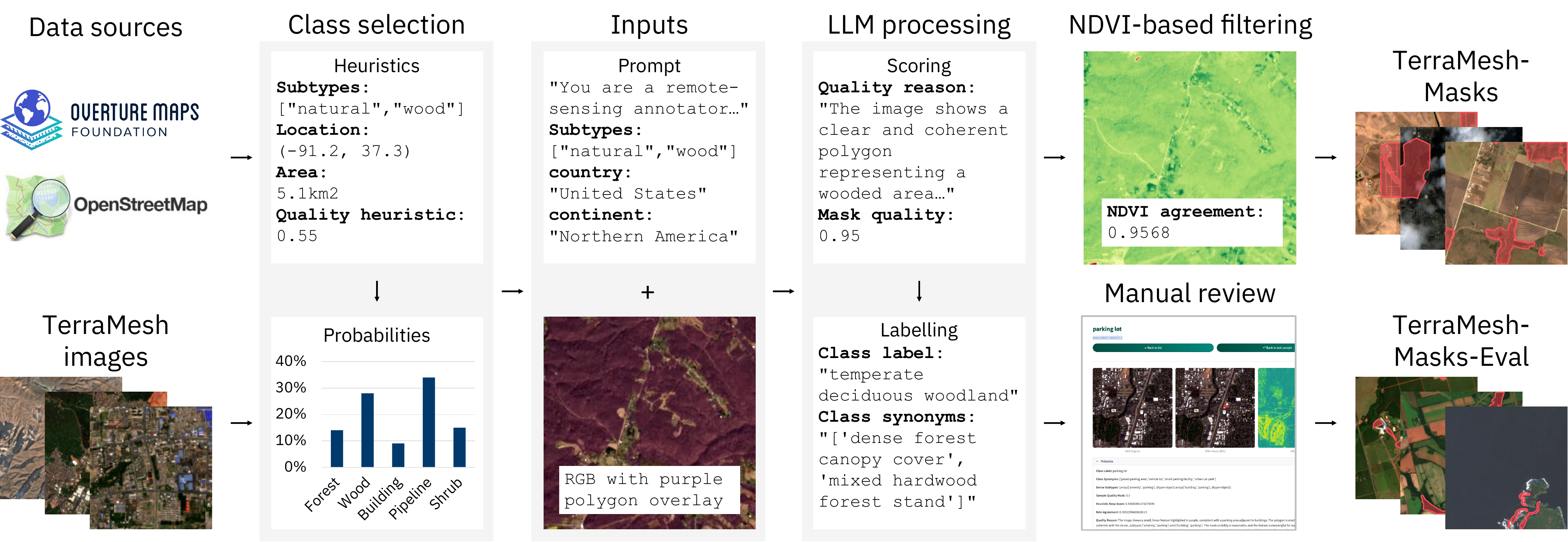}
    \caption{Our dataset creation pipeline pairs TerraMesh samples with Overture and OpenStreetMap annotations. For each sample, we compute heuristics based on subtype, location, polygon size, and previous LLM quality scores. The sampled annotation is processed by Mistral-Large-3 which first scores the annotation quality and then generates the class label and synonyms. The annotations are further subsampled with an NDVI-based filtering for pretraining. We tasked experts to manually review a subset of the dataset to obtain a high-quality evaluation dataset.}
    \label{fig:dataset_pipeline}
\end{figure}

We start from TerraMesh image patches and associated geospatial metadata. We then obtain semantic-mask candidates and refine them into open-vocabulary segmentation annotations. The resulting dataset expands TerraMesh with pixel-level supervision and language annotations suitable for segmentation. The pipeline integrates two complementary geographic data sources to extract candidates for semantic masks. 
\textbf{Overture Maps} provides a land-cover-focused base layer with hundreds of standardized semantic annotations that cover all global regions. A challenge is its semi-automated annotation approach, which can lead to imprecise or incorrect annotations.
\textbf{OpenStreetMap (OSM)} complements Overture tags and represents a rich collection of volunteer-contributed geographic features, from which we extract a broad set of tag categories including \texttt{natural}, \texttt{highway}, \texttt{building}, \texttt{waterway}, \texttt{amenity}, and related structured and semi-structured tags. Compared to Overture, OSM provides rich long-tail semantic coverage, but its schema is heterogeneous and often noisy from the perspective of segmentation supervision. 
Both sources are spatially filtered against TerraMesh image patches, retaining only geometries intersecting the available satellite observations. 
We summarize the data pipeline in Fig.~\ref{fig:dataset_pipeline} and provide additional descriptions of the LLM processing and filtering within this section and the supplementary material.

\paragraph{Preprocessing and geometric filtering.}
Prior to quality assessment, candidate masks undergo several preprocessing steps. First, area-based filtering retains only candidate masks with sufficient spatial extent, using a default threshold of 1{,}000\,m$^2$. This step removes sub-pixel features that are unlikely to be detectable at Sentinel-2 resolution. Second, geometry dissolution merges adjacent or overlapping features with identical semantic labels to create more contiguous regions, reduce fragmentation, and improve spatial coherence. Third, geographic contextualization enriches each mask with country and continent metadata.

\paragraph{Pre-filtering and candidate selection.}
The pipeline employs a hierarchical filtering strategy that combines heuristic pre-filters with LLM-based semantic validation.
At the scale of our raw candidate pool, full LLM validation of every mask is computationally infeasible as the dataset generation starts from roughly 93 million candidate masks. We, therefore, use heuristic pre-filtering not only as a quality-control mechanism, but also as a practical necessity to reduce the candidate set to a tractable subset for large-scale semantic annotation via an LLM. Thus, the filtering stage is important for both efficiency and dataset quality. For example, a naive strategy that submits all raw candidates to an LLM would spend most of the annotation budget on masks that are either clearly implausible, visually unsupported, too small to be useful at Sentinel-2 resolution, or massively overrepresented. Instead, our prefilter concentrates LLM compute on candidates that are more likely to yield valid segmentation supervision while still preserving geographic and semantic diversity. 
After pre-filtering, we sample up to three candidate masks per image using weighted random sampling. The weights are computed by combining the same three heuristics used in filtering, specifically mask area size, previous LLM quality scores of the same annotation tag, and the number of previously selected masks of this tag. The sampling score is composed as follows: $p = p_{\mathrm{freq}} \cdot p_{\mathrm{spec}} \cdot p_{\mathrm{class}}$, where $p_{\mathrm{freq}}$ blends a global and an H3-local (around 100\,km$^2$ areas) quality prior for the annotation tag, $p_{\mathrm{spec}}$ is the NDVI agreement score, and $p_{\mathrm{class}}$ downweights already overrepresented tags. The NDVI agreement score measures spectral separability by comparing the NDVI statistics inside and outside the candidate polygon, so a mask whose interior is spectrally distinct from its surroundings scores higher. The heuristics balance compute efficiency, label exploration, and dataset diversity, which we further outline with full definitions of all terms in the supplementary material.
Our filtering discards 43.5 million candidate masks (46.4\%) and after sample selection reduces the dataset size required to be processed by the LLM to 20.17 million samples (21.5\% of the raw pool). Among those candidates, 16.07 million are ultimately accepted by the LLM as we detail in the next section. That corresponds to an acceptance rate of 79.7\%. We interpret this high post-filter acceptance rate as an indication that the heuristic stage removes a large fraction of obviously poor candidates while preserving a high-yield subset for downstream semantic annotation.

\paragraph{LLM-based validation and labeling.}
Following heuristic pre-filtering, candidate masks undergo LLM-based validation to assess semantic consistency. 
After evaluating the quality and openness of several multimodal LLMs, we selected Mistral-Large-3~\cite{mistrallarge3} for semantic validation and labeling.
The LLM evaluates the visual coherence between the satellite imagery and the provided mask, generating both quality assessments and class labels. For each candidate mask, the LLM receives a structured input containing the satellite imagery as RGB with the candidate polygon overlaid, geometric metadata (including area in m$^2$, country, and continent), the provisioned \texttt{subtypes} tags from OSM or Overture, and raw heuristic quality scores. The prompt casts the model as a conservative annotator, instructs it to treat \texttt{subtypes} as the primary signal for the concept while using the image to detect clear contradictions, penalizes ambiguity, and rejects concepts that are administrative rather than visually grounded, such as legal borders. The model outputs an explanation of visual coherence between the polygon mask and the image patch alongside a quality score (between 0 and 1) as well as a refined natural-language class label synthesizing tag metadata and visual content. Finally, the model provides synonyms as alternative semantic descriptions. We constrain generation to a fixed JSON schema and accept every mask scoring above a cut-off of 0.5. Validation and labeling required around 4{,}000 NVIDIA H200 GPU hours for 20M sampled masks. We provide a longer explaination of the approach in the supplementary material.

\paragraph{Scale and diversity.}
The central value of TerraMesh-Masks is not only its scale, but also the breadth of its vocabulary. We observe more than 44,000 unique class labels for over 16 million labeled masks, together with over 537,000 unique synonym strings. The most frequent labels include broad land-cover concepts such as \textit{shrubland}, \textit{forest}, \textit{grassland}, \textit{barren land}, and \textit{river}. The long tail contains more diverse semantic concepts such as \textit{coral reef}, \textit{golf course}, and \textit{airport taxiway}. The combination of large-scale supervision and rich lexical variation makes TerraMesh-Masks particularly suitable for open-vocabulary segmentation in EO. 

\paragraph{Human-reviewed evaluation set.}
Beyond the large automatically constructed corpus, we curate a human-reviewed evaluation benchmark from the validation split. Four expert annotators provided over 4200 reviews, with each mask being labeled as \textit{valid}, \textit{unsure}, or \textit{invalid}. Using redundant labeling, this process covered 2{,}300 candidate masks. For the evaluation set, we select any mask with at least two \textit{valid} reviews and no \textit{invalid} review. The pairwise agreement rate between two reviewers is between 62\% and 83\% (when ignoring \textit{unsure}). We filter duplicates of the same image by selecting the more infrequent class label, and limit each class label to at most 10 samples to increase diversity. This results in 523 high-quality masks with 207 unique class labels, with examples shown in Fig.~\ref{fig:dataset-overview}. We manually group the selected samples into 14 semantic clusters, used also in Fig.~\ref{fig:model_results}, to better interpret the strengths and weaknesses of OVS models.
For better usability with machine learning models, we center crop the samples to $256 \times 256$ pixels. To avoid data leakage, we remove the selected masks and all spatially overlapping samples from TerraMesh-Masks.

The manual review provides an upper bound on how noisy the automatically constructed corpus is:
Across the 4{,}211 individual reviews, 44\% of verdicts are \textit{valid}, 29\% are \textit{invalid}, and 28\% are \textit{unsure}.
Reasons for rejection include inaccurate or very small masks as well as masks which cover the full image and are, therefore, not well suited for evaluation. We also exclude wrong class labels and and ambiguous classes which are not detectable at 10\,m resolution, like \textit{government building}.
The invalid rate should not be read directly as the training-set error rate, since a mask can be correct but not verifiable from the image alone, which is unusable for evaluation yet still an acceptable training target. Furthermore, masks flagged as invalid are often partially wrong with many correctly labeled pixels (see examples in Fig.~\ref{fig:bad_annotations} of the supplementary material). 

\paragraph{Evaluation protocol.}
Using the expert-reviewed validation samples, we propose TerraMesh‑Masks‑Eval, a subset comprising 523 samples and more than 200 semantic concepts, available at \url{https://huggingface.co/datasets/ibm-esa-geospatial/TerraMesh-Masks-Eval}. For each image–query pair, models must generate a binary mask. We measure performance as Intersection‑over‑Union and F1 score, computed at sample level and aggregated across all samples. To expose systematic patterns in performance, results are additionally reported by high‑level semantic groups such as infrastructure, land cover, and transportation. No restrictions are placed on model architecture or training strategy. The benchmark inherently captures multiple degrees of vocabulary openness. These include queries that are present during training as well as synonyms and lexical variants. We provide the evaluation queries as part of TerraMesh‑Masks‑Eval, ensuring reproducibility and consistent comparison across methods. 

\section{TerraMind-OVS}



\begin{figure}[!b]
  \centering
  \vspace{-5mm}
  \includegraphics[width=0.8\linewidth]{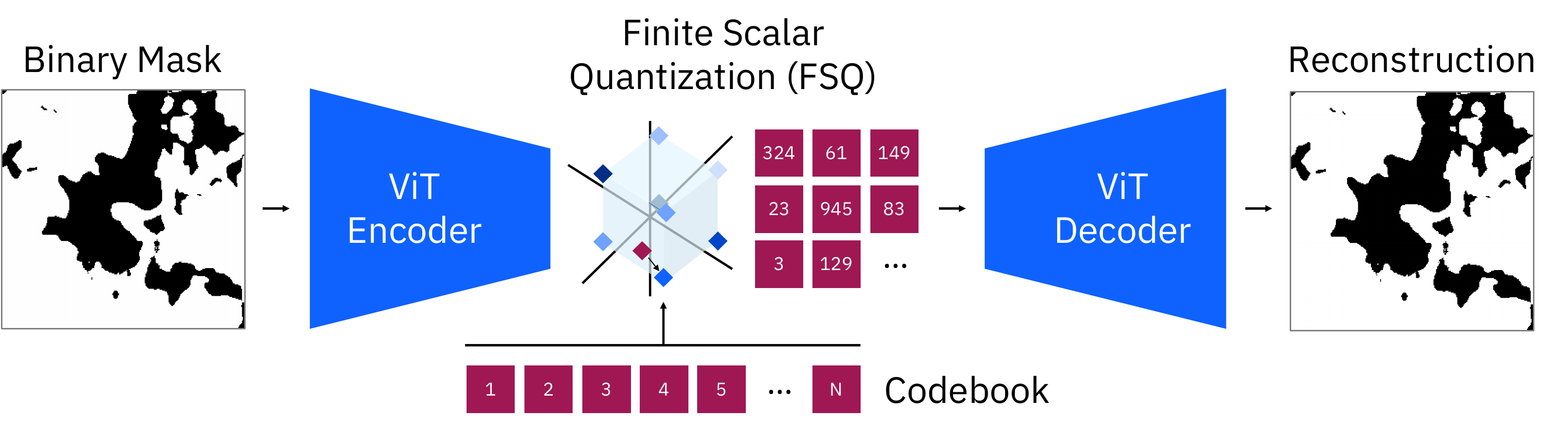}
  \caption{Tokenizer architecture for discretizing binary segmentation masks into tokens. The tokens are later used as training targets of TerraMind-OVS, while the ViT decoder is used during inference.}
  \label{fig:tokenizer}
\end{figure}
\begin{figure}[h]
  \centering
  \includegraphics[width=\linewidth]{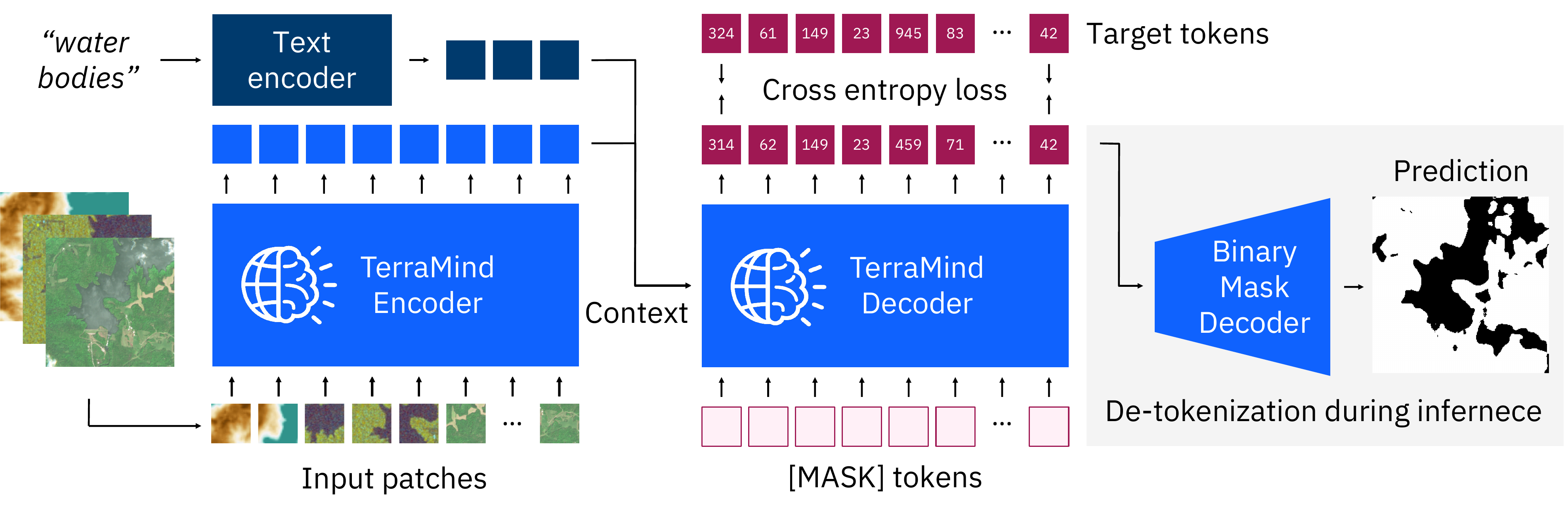}
  \caption{Overview of our model architecture. Multimodal inputs are patchified and encoded with TerraMind. The text query is encoded using a frozen CLIP text encoder followed by an MLP. Both embeddings are used as context by the TerraMind decoder to predict the binary mask tokens. During inference, the predicted tokens are de-tokenized to generate binary segmentation masks.}
  \label{fig:model-architecture}
\end{figure}

We build on top of TerraMind~\cite{jakubik2025terramind} and adapt it to open-vocabulary binary segmentation. We deliberately employ minimal architectural changes in order to demonstrate that performance gains can largely be attributed to our pretraining dataset, TerraMesh-Masks. 
TerraMind was trained on several modalities during pretraining and previous work demonstrated simple adaptation to new modalities with continuous pretraining. Therefore, we treat open-vocabulary binary masks as an additional modality.
TerraMind is trained in a discrete token-prediction paradigm: the decoder predicts token IDs, the encoder-decoder architecture is optimized with via a cross-entropy loss. That structure makes it natural to cast segmentation as predicting a sequence of binary mask tokens, as previously done in literature~\cite{wang2025himtok}. We present an overview of the binary mask tokenizer and TerraMind-OVS in Fig.~\ref{fig:tokenizer} and Fig.~\ref{fig:model-architecture}. 
We continuously pretrain TerraMind's existing multimodal encoder-decoder and primarily specialize the language conditioning as well as the decoding to the mask-prediction setting. 
For language conditioning, we encode the class query with a CLIP text encoder (OpenAI ViT-B-32) and project the resulting embedding into TerraMind's latent space using a lightweight projection head. The 512-dimensional text embedding is passed through a two-layer MLP with GELU activation and hidden size 1024, and then mapped to $5 \times d$ outputs, where $d$ is TerraMind's latent token dimension. The resulting vector is reshaped into five learned query tokens that are concatenated to the decoder context, allowing the model to condition its prediction on the prompt. 

TerraMind-OVS predicts a sequence of 196 mask tokens as output corresponding to a $14 \times 14$ spatial token grid. We use a seperately trained binary-mask tokenizer to discretize each segmentation mask into a grid of discrete tokens, following the setting from TerraMind's Land-Use-Land-Cover tokenizer. In line with the general TerraMind framework, this tokenizer later also allows us to decode predicted tokens back to pixel space. 

\section{Experimental setup}


\paragraph{Training setup.}
We continuously pretrain TerraMind variants \textit{small}, \textit{base}, and \textit{large}. 
Training samples from TerraMesh-Masks are processed through a filtered WebDataset pipeline. To mitigate class imbalance, we apply class-dependent repeat-factor resampling during training. We also filter out masks with invalid geometry, such as masks that are too small or span the full frame. Images are cropped to $224 \times 224$ pixels.

Across our training setups, models are trained with inputs from Sentinel-1 RTC, Sentinel-1 GRD, Sentinel-2 L2A, Sentinel-2 RGB, DEM, and a text query, and generally only predict tokens of the binary semantic mask on a patch size of $16$. We train two model versions, one with all modalities and one limited to S2 L2A and RGB, labeled with S2 in the model name.
The text query is built from the class label associated with each mask. 
We use the MS-CLIP prompt templates~\cite{marimo2025beyond} to adapt the embeddings to satellite data.
When synonyms are available, we sample between the main label and its variants, with the main label sampled with a factor $f=3$ relative to any synonym during training. 
This exposes the model to linguistic variations while keeping training anchored to the dataset taxonomy.

The supervision signal is cross-entropy on the target mask tokens. We train TerraMind-OVS-S with a global batch size of 1024 and a learning rate of 5e-4, while base and large models use a batch size of 512 and a learning rate of 1e-4. All runs use an AdamW optimizer with a cosine decay scheduler and 1 million samples for warmup. Differrent from TerraMind pretraining, we do not apply any target masking and each sample includes 196 target tokens for a full binary mask. With 9M samples in TerraMesh, we contiuously pretrain for 15B tokens (3\% of the original training). Training for TerraMind-OVS-L required around 250 NVIDIA A100 80GB GPU hours.

\paragraph{Benchmark setup.}
While TerraMind-OVS is trained on $224 \times 224$ inputs, we pass the full $256 \times 256$ images without resizing during evaluation. For multiclass datasets, we run the model per class and stack the logits before appling the argmax step.
For benchmark models, we resize the images to the default image resolution such as $224 \times 224$ pixels for DINO Soars, $512 \times 512$ for DINOv3.txt, and $1008 \times 1008$ for SAM-3. We use a confidence threshold of 0.1 for SAM-3 and 0.5 for RemoteSAM, which optimizes the models' mIoU scores. DINOv3.txt uses the ImageNet prompt templates, while we also tested MS-CLIP templates, which led to lower performance. For DINOv3.txt, DINO Soars, and SegEarth-OV, we use a \textit{background} label for the negative class after testing several candidates.
For GeoPixel, we follow the default settings of reshaping the image to $1680 \times 1680$ pixel and processing it in 9 patches. We use \textit{Could you provide a segmentation mask for {label} in this image?} as a prompt following the authors' examples.





\section{Results}


\begin{figure}[tbh]
    \centering
    \includegraphics[width=\linewidth]{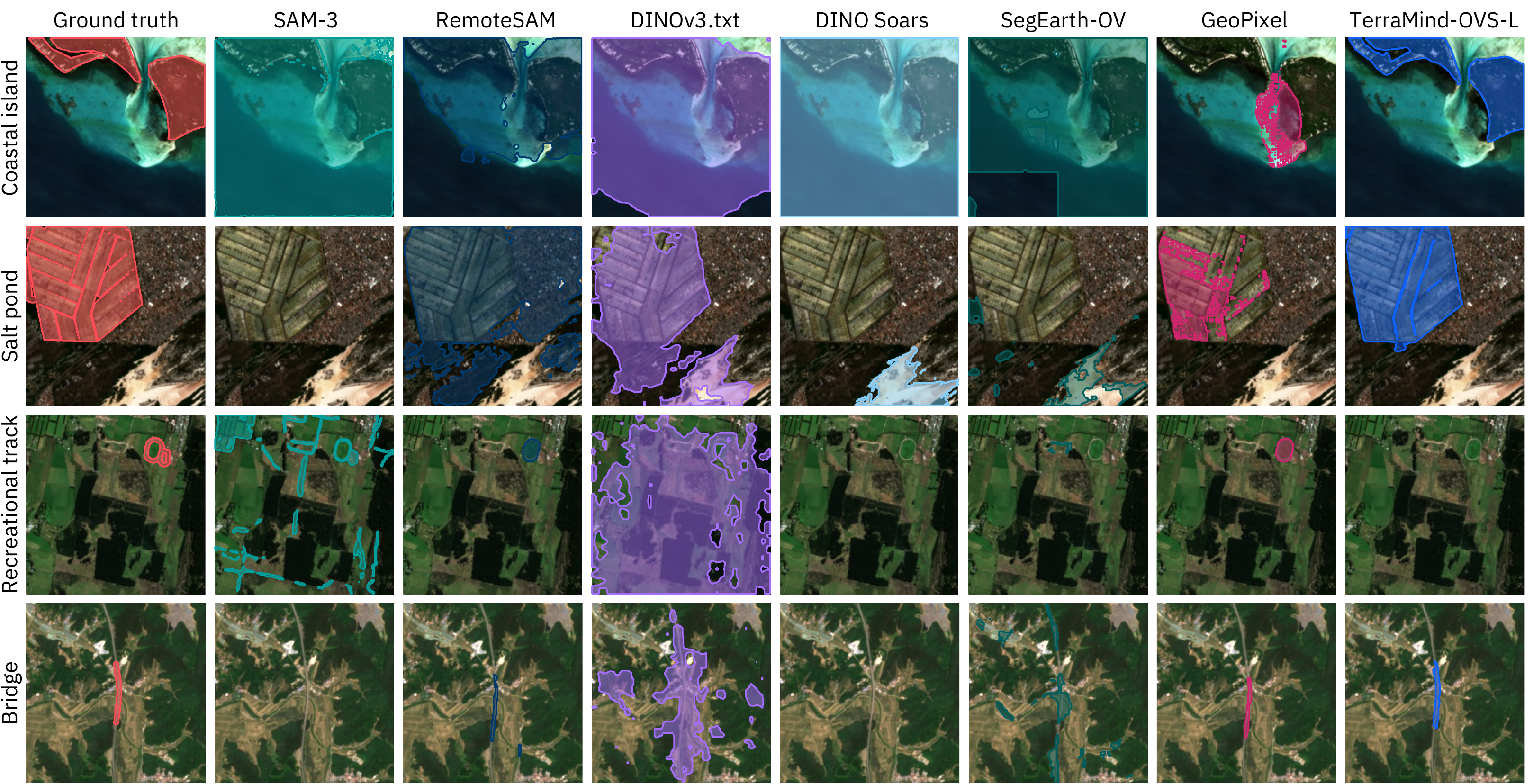}
    \caption{Qualitative comparison on examples from TerraMesh-Masks-Eval. TerraMind performs well on larger land-cover related labels, while GeoPixel is better in spotting objects. Other models lean towards overprediction or no prediction.}
    \label{fig:predictions}
\end{figure}

\paragraph{Evaluation performance.}
We summarize the performance of TerraMind-OVS and baselines in Tab.~\ref{tab:tm-masks-eval-results} and report Student-t 95\% confidence intervals over the 523 evaluation samples. 
We compare against generalist open-vocabulary models (SAM-3~\cite{carion2025sam}, DINOv3.txt~\cite{simeoni2025dinov3}) as well as four recent EO-specific ones (RemoteSAM~\cite{yao2025remotesam}, DINO Soars~\cite{faulkenberry2026dinosoars}, SegEarth-OV~\cite{li2025segearthov}, GeoPixel~\cite{shabbir2025geopixel}).
Overall, we observe that TerraMind-OVS outperforms all baselines on EO-specific data across most metrics and model sizes.
TerraMind-OVS-L improves over SAM-3 and DINOv3.txt by 479\% and 53\%, respectively, measured in relative mIoU improvement. 
The EO-specific baselines are considerably stronger, with SegEarth-OV reaching the best IoU$_{pos}$ with 21.59\%, yet TerraMind-OVS-L still improves on it by 13.59pp IoU$_{pos}$. GeoPixel has the second-best mIoU score, 14.29pp below TerraMind.
That emphasizes the relevance and impact of TerraMesh-Masks as a domain-specific pretraining dataset. 
Fig.~\ref{fig:predictions} illustrates zero-shot predictions of TerraMind-OVS models and baselines at inference time. 
Most baseline models lean towards overprediction, which also results in a high average recall.
\begin{wrapfigure}[19]{r}{0.40\textwidth}
  \vspace{-2.2mm}
  \begin{center}
    \includegraphics[width=0.37\textwidth]{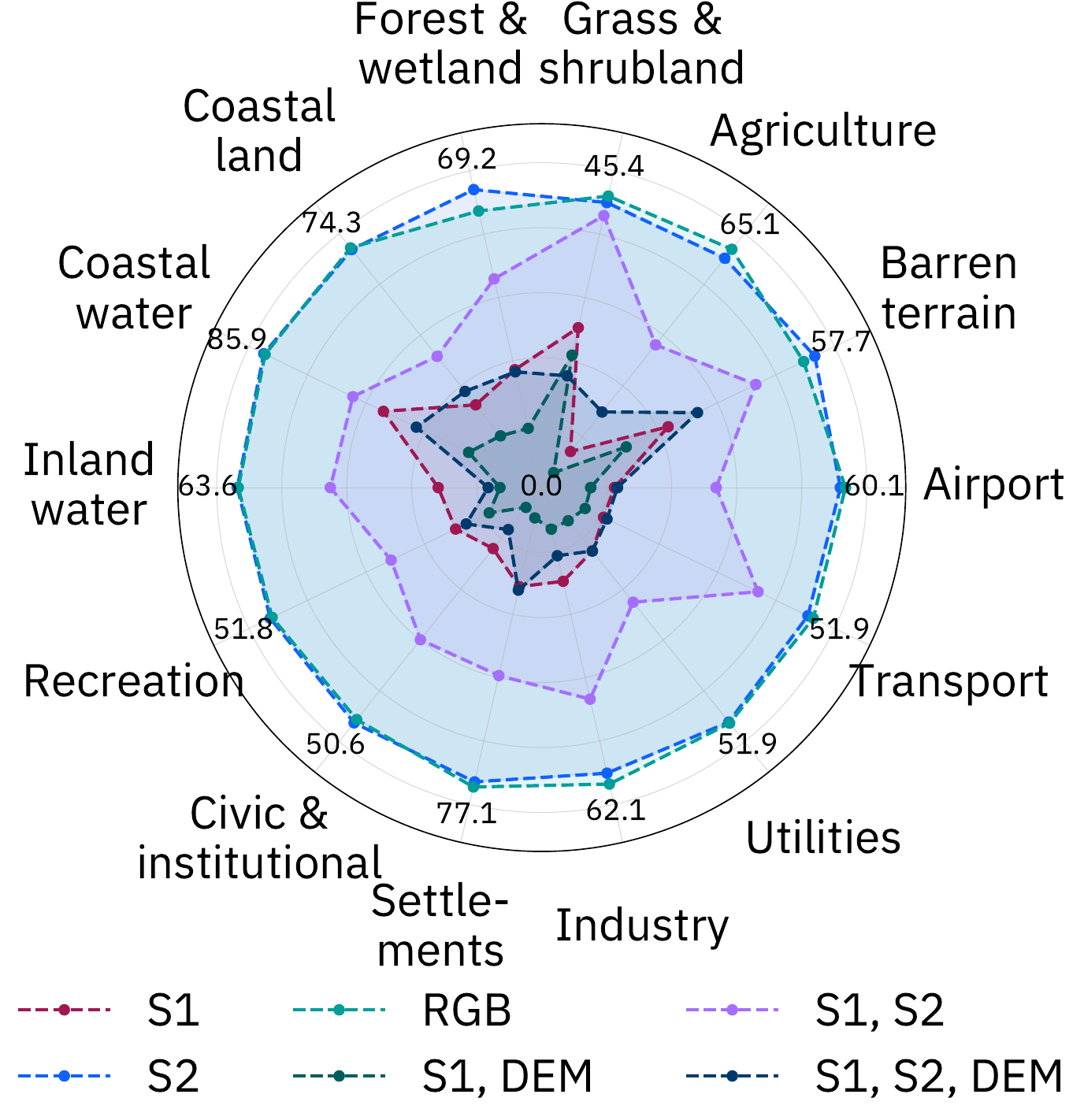}
  \end{center}
  \caption{Eval mIoU results per semantic group for TerraMind-OVS-B with different modality inputs. S2 and RGB perform compareable, while S1 and DEM significalty reduce the performance.}
  \label{fig:results_modalities}
\end{wrapfigure}
We report average inference times on a single A100: SAM-3 requires 310\,ms per sample, DINO Soars 319\,ms, SegEarth-OV 488\,ms, RemoteSAM 524\,ms, and DINOv3.txt 432\,ms, while GeoPixel is significantly slower with 6{,}298\,ms. In comparison, the TerraMind-OVS architecture is more efficient and requires between 52 and 73\,ms, mainly limited by data loading speed rather than inference. 

\begin{table}[h]
    \small
    \caption{TerraMesh-Masks-Eval results with Student-t 95\% confidence intervals, each metric averaged over the 523 per-sample results. Some TerraMind versions (TM-OVS) are trained on multimodal data that includes S1, S2, RGB, and DEM. Best results are in bold, second-best results are underlined.}
    \label{tab:main_results}
    \setlength\tabcolsep{2pt}
    \centering
    \begin{tabularx}{\linewidth}{lrccCCCCC}
    \toprule
    Model & Param. & Training & Input & IoU$_{pos}$ & mIoU & F1 & Precision & Recall \\
    \midrule
    SAM-3 & 841M & RGB & RGB & 11.33 {\tiny$\pm$ 2.30} & 10.97 {\tiny$\pm$ 1.39} & 13.32 {\tiny$\pm$ 2.54} & 12.86 {\tiny$\pm$ 2.58} & 20.97 {\tiny$\pm$ 3.33} \\
    RemoteSAM & 242M & RGB & RGB & 14.79 {\tiny$\pm$ 2.13} & 30.42 {\tiny$\pm$ 1.81} & 19.62 {\tiny$\pm$ 2.48} & 20.60 {\tiny$\pm$ 2.84} & \underline{50.19} {\tiny$\pm$ 3.49} \\
    DINOv3.txt & 867M & RGB & RGB & 16.29 {\tiny$\pm$ 1.87} & 41.40 {\tiny$\pm$ 1.42} & 23.14 {\tiny$\pm$ 2.28} & 24.70 {\tiny$\pm$ 2.72} & \textbf{51.53} {\tiny$\pm$ 3.50} \\
    DINO Soars & 869M & RGB & RGB & 19.49 {\tiny$\pm$ 2.55} & 44.29 {\tiny$\pm$ 1.93} & 24.57 {\tiny$\pm$ 2.83} & 23.65 {\tiny$\pm$ 2.92} & 48.43 {\tiny$\pm$ 3.98} \\
    SegEarth-OV & 150M & RGB & RGB & 21.59 {\tiny$\pm$ 2.47} & 48.38 {\tiny$\pm$ 1.53} & 28.01 {\tiny$\pm$ 2.77} & 30.81 {\tiny$\pm$ 3.10} & 45.34 {\tiny$\pm$ 3.51} \\
    GeoPixel & 11.3B & RGB & RGB & 15.06 {\tiny$\pm$ 1.99} & 49.25 {\tiny$\pm$ 1.58} & 20.62 {\tiny$\pm$ 2.41} & 40.27 {\tiny$\pm$ 3.69} & 20.15 {\tiny$\pm$ 2.57} \\
    \midrule
    TM-OVS-S & 248M & Multimodal & RGB & 26.52 {\tiny$\pm$ 3.28} & 58.76 {\tiny$\pm$ 1.71} & 29.75 {\tiny$\pm$ 3.50} & 33.80 {\tiny$\pm$ 3.72} & 30.15 {\tiny$\pm$ 3.54} \\
    TM-OVS-S & 247M & S2, RGB & RGB & 27.37 {\tiny$\pm$ 3.27} & 59.08 {\tiny$\pm$ 1.71} & 30.88 {\tiny$\pm$ 3.51} & 34.24 {\tiny$\pm$ 3.72} & 32.42 {\tiny$\pm$ 3.62} \\
    TM-OVS-B & 421M & Multimodal & RGB & 32.28 {\tiny$\pm$ 3.42} & 62.00 {\tiny$\pm$ 1.78} & 36.18 {\tiny$\pm$ 3.64} & 40.12 {\tiny$\pm$ 3.80} & 37.79 {\tiny$\pm$ 3.74} \\
    TM-OVS-B & 420M & S2, RGB & RGB & 32.18 {\tiny$\pm$ 3.40} & 61.67 {\tiny$\pm$ 1.83} & 36.20 {\tiny$\pm$ 3.62} & 40.79 {\tiny$\pm$ 3.82} & 37.90 {\tiny$\pm$ 3.73} \\
    TM-OVS-L & 935M & Multimodal & RGB & 34.87 {\tiny$\pm$ 3.35} & 62.90 {\tiny$\pm$ 1.90} & 39.75 {\tiny$\pm$ 3.59} & \underline{45.32} {\tiny$\pm$ 3.80} & 40.15 {\tiny$\pm$ 3.66} \\
    TM-OVS-L & 934M & S2, RGB & RGB & 34.96 {\tiny$\pm$ 3.36} & \underline{63.27} {\tiny$\pm$ 1.76} & \underline{39.83} {\tiny$\pm$ 3.59} & 44.94 {\tiny$\pm$ 3.79} & 40.61 {\tiny$\pm$ 3.66} \\
    \midrule
    TM-OVS-S & 248M & Multimodal & S2 & 26.34 {\tiny$\pm$ 3.26} & 58.49 {\tiny$\pm$ 1.71} & 29.63 {\tiny$\pm$ 3.48} & 34.86 {\tiny$\pm$ 3.76} & 29.97 {\tiny$\pm$ 3.53} \\
    TM-OVS-S & 247M & S2, RGB & S2 & 27.28 {\tiny$\pm$ 3.28} & 58.98 {\tiny$\pm$ 1.72} & 30.75 {\tiny$\pm$ 3.50} & 33.91 {\tiny$\pm$ 3.71} & 32.43 {\tiny$\pm$ 3.63} \\
    TM-OVS-B & 421M & Multimodal & S2 & 32.19 {\tiny$\pm$ 3.42} & 62.04 {\tiny$\pm$ 1.77} & 36.08 {\tiny$\pm$ 3.64} & 39.86 {\tiny$\pm$ 3.81} & 37.54 {\tiny$\pm$ 3.74} \\
    TM-OVS-B & 420M & S2, RGB & S2 & 32.44 {\tiny$\pm$ 3.41} & 62.04 {\tiny$\pm$ 1.82} & 36.47 {\tiny$\pm$ 3.62} & 40.73 {\tiny$\pm$ 3.81} & 38.25 {\tiny$\pm$ 3.73} \\
    TM-OVS-L & 935M & Multimodal & S2 & \underline{35.09} {\tiny$\pm$ 3.35} & 62.95 {\tiny$\pm$ 1.89} & \textbf{40.04} {\tiny$\pm$ 3.59} & \textbf{45.57} {\tiny$\pm$ 3.80} & 40.65 {\tiny$\pm$ 3.66} \\
    TM-OVS-L & 933M & S2, RGB & S2 & \textbf{35.18} {\tiny$\pm$ 3.41} & \textbf{63.54} {\tiny$\pm$ 1.81} & 39.80 {\tiny$\pm$ 3.63} & 44.76 {\tiny$\pm$ 3.82} & 40.97 {\tiny$\pm$ 3.71} \\
    \bottomrule
    \end{tabularx}
    \label{tab:tm-masks-eval-results}
\end{table}

\begin{figure}[hbt]
  \centering
  \includegraphics[width=\textwidth]{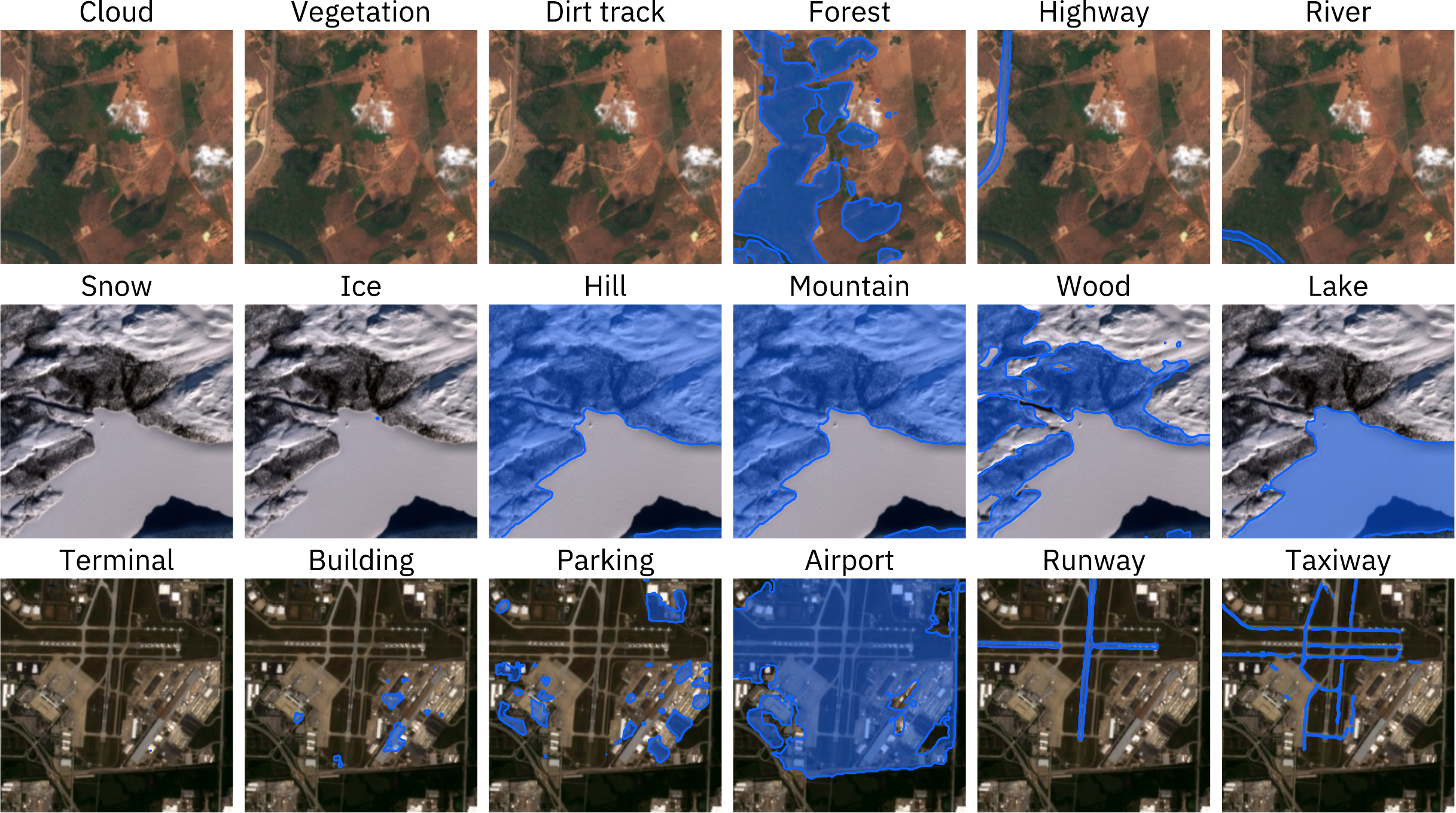}
  \caption{These predictions from TerraMind-OVS-L (S2) show that the model does not have an understanding of \textit{clouds}, \textit{snow}, and other weather patterns as these are not included in the annotations. The model covers a range of concepts, but can still fail for related terms like \textit{dirt track} or \textit{terminal}.}
  \label{fig:failure_cases}
\end{figure}

\paragraph{Failure cases.}
Beyond reporting gains over existing models, we showcase a qualitative assessment in Fig.~\ref{fig:failure_cases} to demonstrate limitations of our model. TerraMind-OVS struggles on classes that are not part of TerraMesh-Masks and, therefore, completely unseen, such as weather patterns (\textit{clouds} and \textit{snow}) and specific terminology that is not covered by the synonyms (e.g., \textit{terminal}).

\paragraph{Effect of multimodal and multispectral inputs.}

We ablate the effect of multimodal and multispectral inputs to TerraMind-OVS in Tab.~\ref{tab:tm-masks-eval-results} and in Fig.~\ref{fig:results_modalities}. Overall, we observe top performances for using multispectral S2 input and a generally strong performance with RGB inputs that is likely related to the fact that the vast majority of objects in OpenStreetMap and Overture Maps are visible in RGB. We observe particular benefits of multispectral inputs on groups like \textit{Forest \& Wetland} as well as \textit{Barren terrain}. Contrary to our expectations, multimodal input during inference does not further enhance the predictions of TerraMind-OVS. This points to a need for more sophisticated multimodal architectural designs for multimodal OVS.

\paragraph{Multiclass segmentation.}

In Tab.~\ref{tab:multilabel}, we test TerraMind-OVS in more challenging multi-label segmentation and out-of-distribution scenarios. Our models outperform generalist models by up to 7.78pp mIoU on reBEN-7k~\cite{clasen2025reben} and perform competitively on out‑of‑distribution grayscale SAR imagery from the CaFFe dataset~\cite{gourmelon2023caffe}. Since CaFFe does not provide true SAR amplitude data, we replicate the grayscale channel to construct a pseudo‑RGB input for all models, ensuring fair comparison (see supplementary material). Interestingly, we generally observe less scaling behavior in model size for the multi-label evaluations compared to the experiments on TerraMesh-Masks-Eval. We largely attribute this to how challenging the underlying task is.

\begin{table}[h]
\centering
\small
\caption{Open-vocabulary segmentation performance across two multi-label datasets: CaFFe \citep{gourmelon2023caffe} and reBEN-7k \citep{clasen2025reben,marti2025fine}. Thresholds for SAM-3 are optimized over the test set for maximum performance.}
\label{tab:multilabel}
\begin{tabular}{lcccccc}
\toprule
{Model} 
& \multicolumn{2}{c}{{CaFFe (OOD)}} 
& \multicolumn{2}{c}{{reBEN-7k (Test split)}} 
& \multicolumn{2}{c}{{reBEN-7k (OOD split)}} \\
\cmidrule(lr){2-3} \cmidrule(lr){4-5} \cmidrule(lr){6-7}
& mIoU (\%) & F1 (\%) & mIoU (\%) & F1 (\%) & mIoU (\%) & F1 (\%) \\
\midrule

DINOv3.txt 
& \textbf{15.32} & \underline{22.64}
& 3.52 & 5.20
& 3.07 & 4.60 \\

SAM-3 
& 1.00 & 1.24 
& 7.30 & 1.00 
& 8.76 & 1.03 \\\midrule 

TerraMind-OVS-S
& \underline{14.57} & \textbf{22.82}
& 10.93 & 18.40
& 11.26 & 17.59 \\

TerraMind-OVS-S (S2)
& 14.35 & 22.29
& 11.46 & 19.25
& 14.06 & 21.23  \\

TerraMind-OVS-B
& 9.10 & 14.14
& \textbf{13.75} & \textbf{22.23}
& \textbf{16.54} & \textbf{24.32} \\

TerraMind-OVS-B (S2)
& 12.10 & 19.39
& 12.60 & 20.72
& 14.69 & \underline{21.95} \\

TerraMind-OVS-L
& 8.24 & 12.40
& \underline{13.66} & \underline{21.97}
& 12.17 & 19.86 \\

TerraMind-OVS-L (S2)
& 8.26 & 12.44
& 12.74 & 20.87
& \underline{14.89} & 21.93  \\
\bottomrule
\end{tabular}
\end{table}


\section{Limitations}
\label{sec:limitations}

\textbf{Scope of the vocabulary.}
OpenStreetMap and Overture Maps include persistent map features, so TerraMesh-Masks describes static land cover and infrastructure.
Dynamic phenomena such as clouds, floods, fire, or seasonal change are absent from the vocabulary, and a model trained only on our dataset cannot segment them.
Covering them would require annotations that are time-stamped against the acquisition date of the image.

\textbf{Annotation noise.}
The corpus is constructed automatically and therefore introduces noise.
Wrong class labels, imprecise boundaries, and concepts that are not resolvable at 10\,m all remain in the training data.
Our filtering heuristics reduce but do not eliminate them, as the manual review of the evaluation samples demonstrates.

\textbf{Geographic coverage.}
Both the training corpus and the evaluation set inherit the geographic bias of OpenStreetMap, which is denser in Europe and North America than in much of Africa, Central Asia, and South America.
High latitudes are barely covered, with 13 of the 523 evaluation samples above 60\textdegree{} and none in Antarctica.

\textbf{Size of the evaluation set.}
TerraMesh-Masks-Eval contains 523 expert-reviewed masks, which we report with 95\% confidence intervals throughout.
The intervals are wide enough that it remains challenging to distinguish different TerraMind-OVS variants from each other. The per-group breakdowns rest on as few as ten samples for the rarest groups.
The benchmark is reliable for the large differences between model families that we discuss, and should not be used to rank closely matched checkpoints.

\section{Conclusion}

In this work, we address a critical gap in open-vocabulary segmentation for Earth observation by introducing TerraMesh-Masks, a large-scale multimodal OVS dataset designed to reflect the richness and complexity of satellite data. By extending TerraMind to open-vocabulary segmentation, we demonstrate the value of TerraMesh-Masks and its potential to unlock substantial performance gains. We hope that our dataset, benchmarks, and strong baselines lay the foundation for scalable, modality-aware OVS in EO and enable broader adoption and future research in this field.

\begin{ack}
This work is part of the FAST-EO – CCN2 project funded by the European Space Agency (ESA) $\Phi$-Lab under contract number 4000143501/23/I-DT.
\end{ack}

\bibliographystyle{unsrtnat} 
\newpage
\bibliography{references}

\newpage
\appendix

\begin{center}
\Large\bfseries TerraMesh-Masks\\Supplementary Material
\end{center}

The supplementary material is organized as follows:

\begin{itemize}
    \item We provide an overview of the TerraMesh source dataset in Section \ref{sup:terramesh}.
    \item More information on the dataset and the captioning pipeline is described in Section \ref{sup:dataset}.
    \item Training details are outlined in Section \ref{sec:training_details}.
    \item Lastly, we provide additional quantitative and qualitative results in Section \ref{sup:add_results}.
\end{itemize}

\section{TerraMesh}
\label{sup:terramesh}

\begin{figure}[hb]
    \centering
    \includegraphics[width=\linewidth]{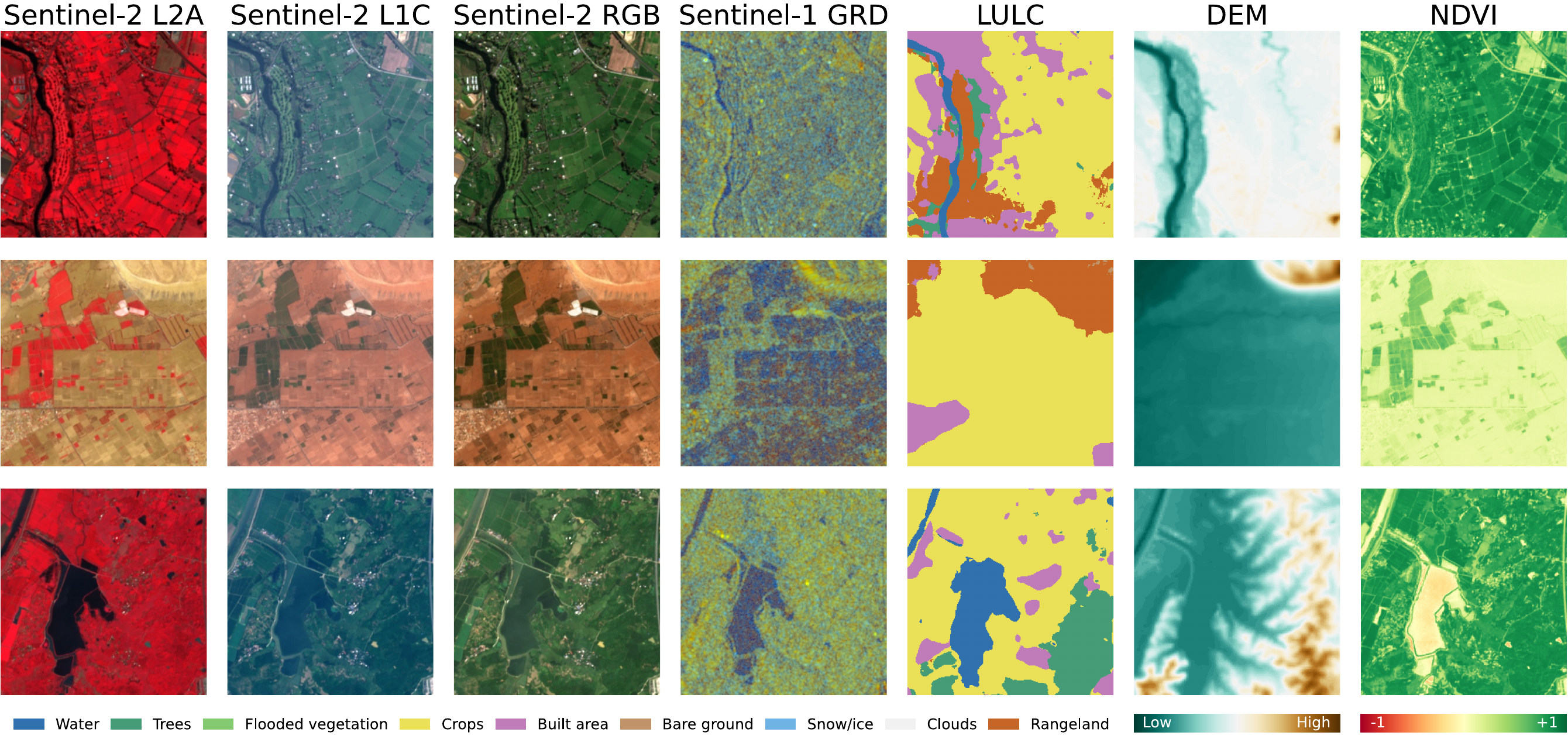}
    \caption{Modality examples in TerraMesh. We exclude NDVI and LULC from training and from the Eval subset as these modalities are related to specific semantic classes and might bias OVS models.}
    \label{fig:modalities}
\end{figure}

\begin{figure}[bht]
    \centering        
    \begin{subfigure}[b]{0.5\linewidth}  
    \centering
    \includegraphics[width=\linewidth]{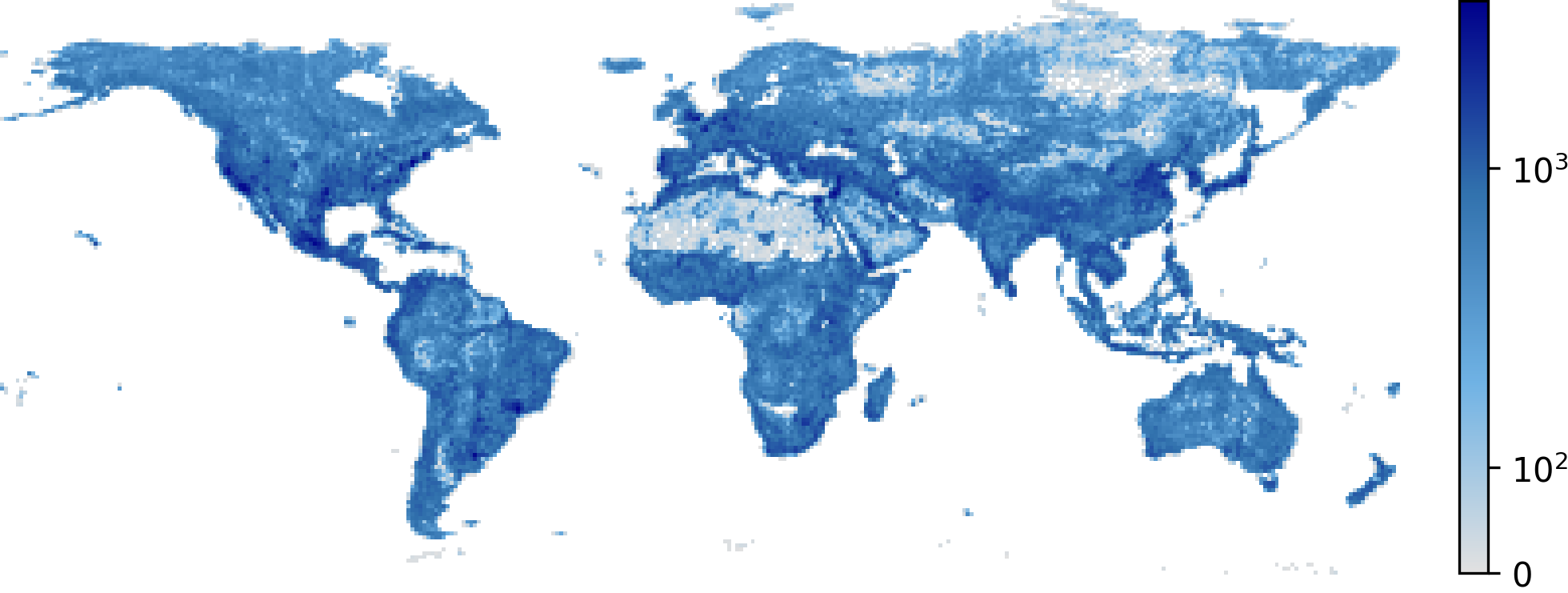}
    \vspace{1mm}
    \label{subfig:spatial}
    \caption{Spatial distribution}
    \end{subfigure}
    \hfill
    \begin{subfigure}[b]{0.48\linewidth}    
    \centering
    \includegraphics[width=\linewidth]{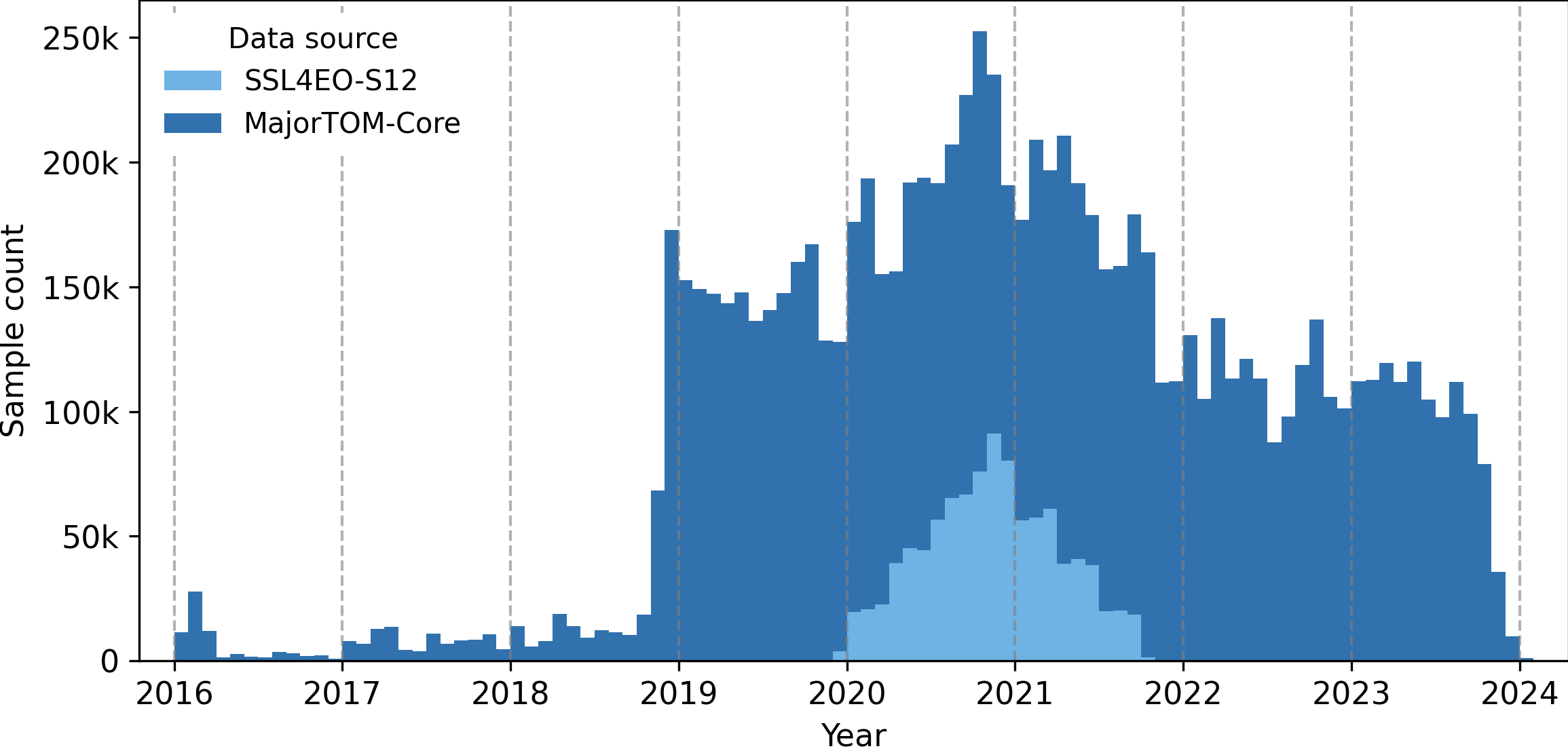}
    \label{subfig:temporal}
    \caption{Temporal distribution}
    \end{subfigure}
    \caption{Spatiotemporal distribution of the TerraMesh source dataset with (a) sample count visualized on a one degree map, and (b) monthly temporal distribution of its subsets using the S2 timestamps.}
    \label{fig:distribution}
\end{figure}

TerraMesh~\cite{blumenstiel2025terramesh} is a large-scale, analysis-ready multimodal EO dataset comprising over 9 million co-registered samples at 10\,m resolution, with a patch size of $264 \times 264$ pixels, designed explicitly for scalable pretraining workflows. Each sample contains a consistent set of spatially aligned modalities, including Sentinel‑2 optical imagery (L1C top-of-atmosphere, L2A surface reflectance, and derived RGB), Sentinel‑1 SAR backscatter (GRD or RTC depending on the subset), NDVI, Copernicus DEM, and ESRI land-use/land-cover annotations, all harmonized to a common grid as shown in Fig.~\ref{fig:modalities}. The dataset provides global coverage across diverse ecosystems and ecoregions as visualized in Fig.~\ref{fig:distribution}.
TerraMesh is constructed by combining near-global single-timestamp data from MajorTOM-Core~\cite{francis2024major} with multi-season urban-centric patches from SSL4EO-S12~v1.1~\cite{blumenstiel2025ssl4eo}. 
Homogeneous regions (e.g., deserts) in MajorTOM are explicitly downsampled. Temporally, acquisitions span eight years (primarily 2019–2023) with S1 and S2 images largely aligned within one week. 

For practical usage, TerraMesh is distributed as shuffled WebDataset tar archives (17\,TB) containing compressed Zarr samples with colocated metadata (e.g., geolocation and timestamps), enabling efficient streaming and data loading on large-scale compute systems.
TerraMesh-Masks uses a similar data structure and natively integrates with the WebDataset structure of TerraMesh as an additional modality. Our dataset requires an additional 1.2\,TB of storage and we adapted the data loading script to image-text pairs.

\section{Dataset details}
\label{sup:dataset}

\begin{figure}[tbh]
    \centering
    \includegraphics[width=0.7\linewidth]{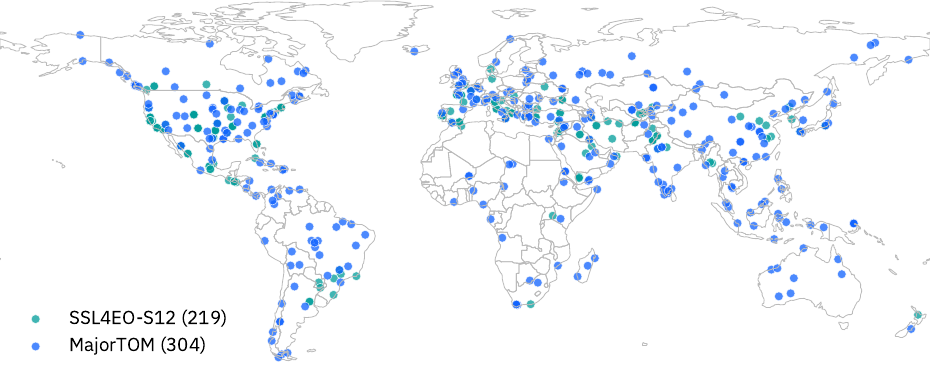}
    \caption{Geographic distribution of the 523 samples in TerraMesh-Masks-Eval, coloured by their source dataset.}
    \label{fig:eval_distribution}
\end{figure}

\begin{figure}[!h]
  \centering
   \includegraphics[width=0.95\textwidth]{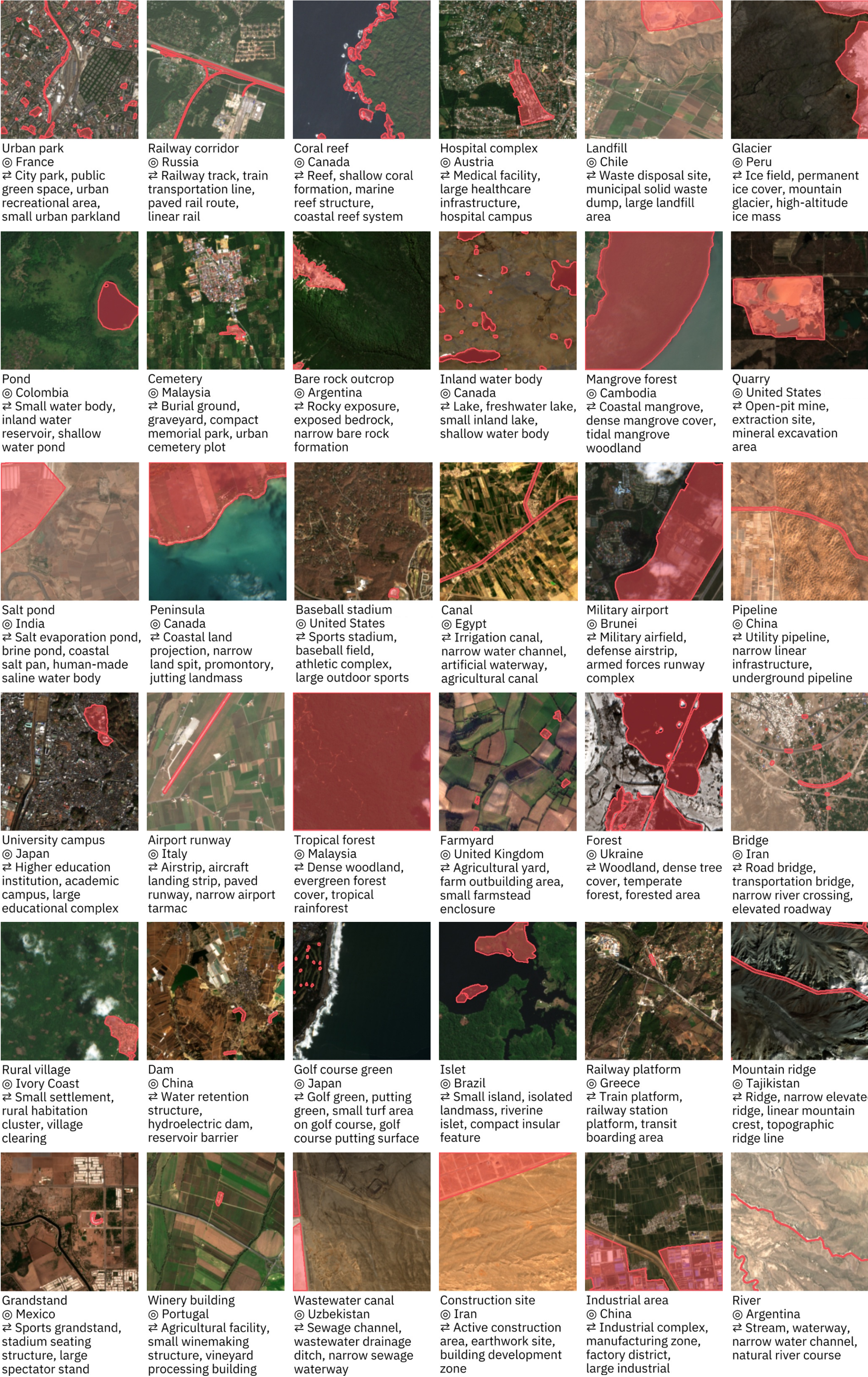}
  \caption{Additional examples from the TerraMesh-Masks-Eval set. Each patch is shown with a polygon mask overlay and its open-vocabulary class label, country, and a synonym set.}
  \label{fig:supp-examples}
\end{figure}

\textit{Evaluation set.}
Fig.~\ref{fig:eval_distribution} shows where the 523 evaluation samples of TerraMesh-Masks-Eval are located.
They span 83 countries, and coverage is densest in North America (126 samples) and Europe (101 samples), while Africa contributes 23 samples in total.
This mirrors the mapping density of OpenStreetMap rather than a sampling choice on our side.
Only 13 samples lie above 60\textdegree{} latitude, since TerraMesh excludes the polar regions.
We provide further examples drawn from the Eval subset in Fig.~\ref{fig:supp-examples} to demonstrate the diversity of semantic concepts. An overview of the groups is provided in Tab.~\ref{tab:groups_stats} with sample counts and labels.

\begin{figure}
    \vspace{-3mm}
    \centering
    \includegraphics[width=\linewidth]{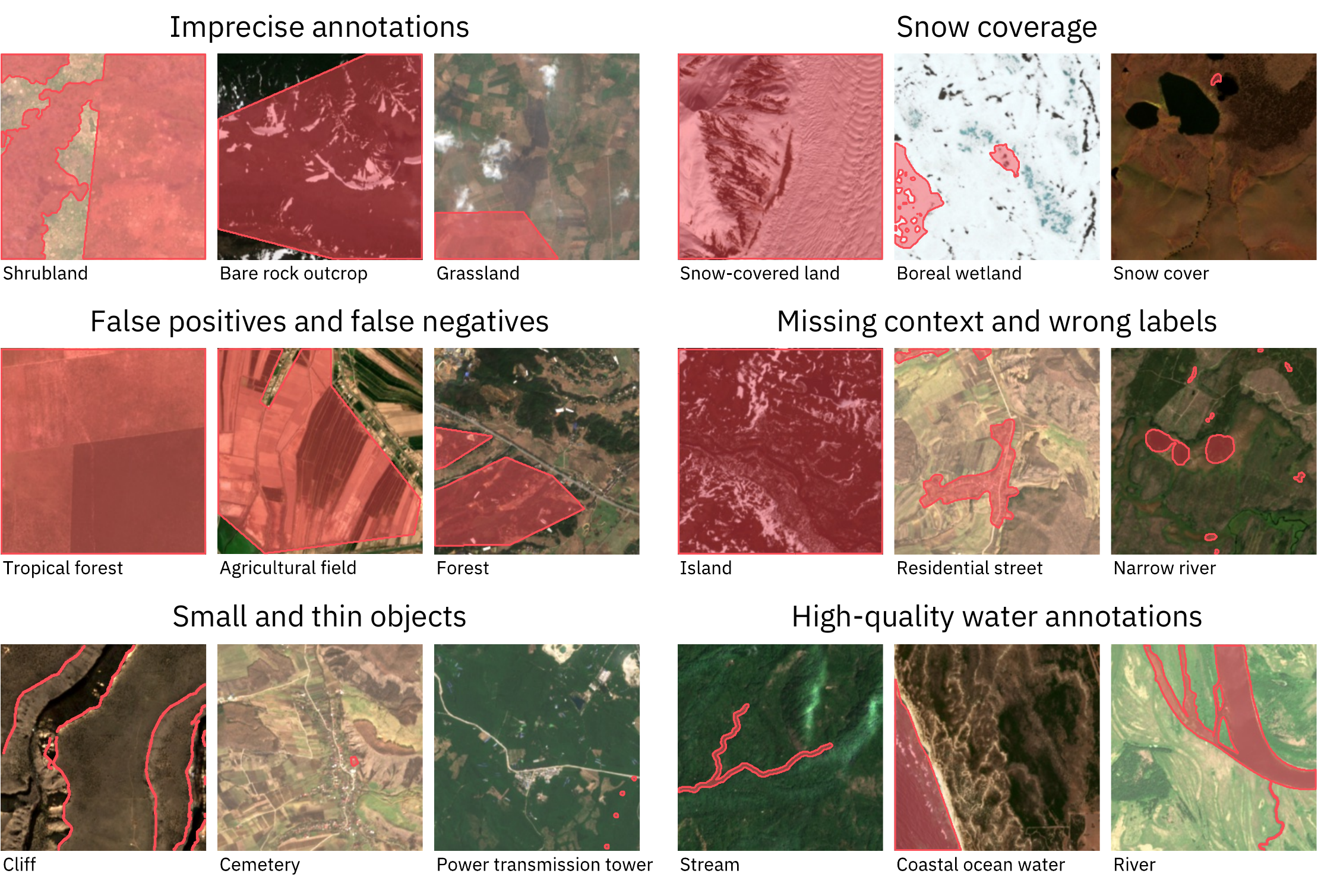}
    \caption{Wrong or challenging annotations. The dataset contains several noisy labels, while water-related classes are often very precise.}
    \label{fig:bad_annotations}
    \vspace{-3mm}
\end{figure}

\begin{figure}
    \vspace{-3mm}
    \centering
    \includegraphics[width=0.7\linewidth]{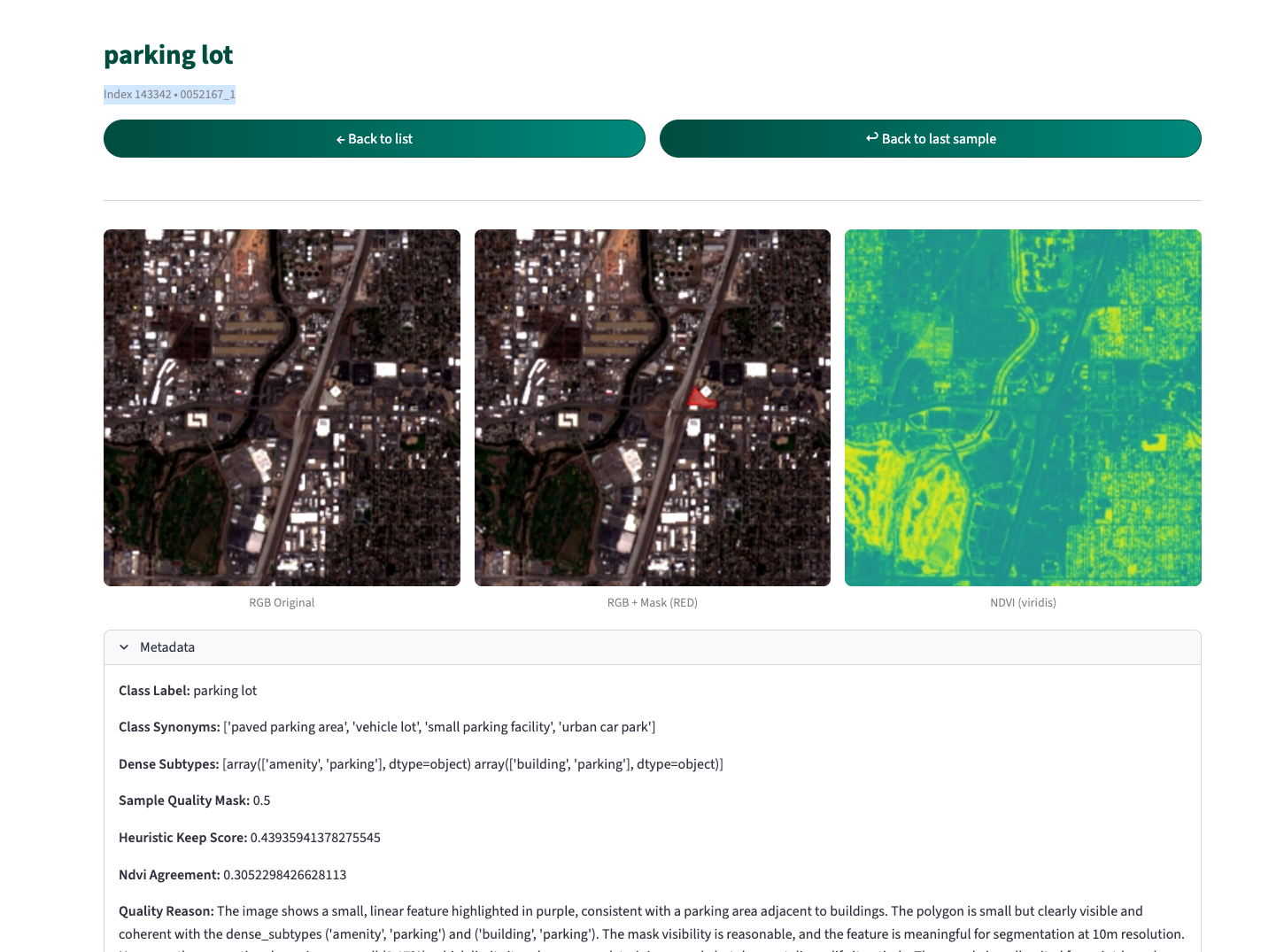}
    \caption{Example of the provided information to the reviewers when creating the TerraMesh-Masks-Eval dataset. Based on this information, reviewers rated the image-mask-text combination as \textit{valid}, \textit{invalid}, or \textit{unsure}.}
    \label{fig:annotation}
    \vspace{-5mm}
\end{figure}

Despite the filtering processes outlined in the following subsections, TerraMesh-Masks still contains several noisy labels as shown in Fig.~\ref{fig:bad_annotations}. We, therefore, created and released the evaluation set based on human reviews. In Fig.~\ref{fig:annotation} we show the information that was provided to the reviewers. If the information was not sufficient to accurately label a mask as \textit{valid} or \textit{invalid}, reviewers could always label them as \textit{unsure}.

\begin{table}[h]
    \vspace{-3mm}
    \footnotesize
    \caption{Overview of the assigned label groups based on manual review and grouping of 200+ class labels in TerraMesh-Masks-Eval. We display the most common class names of each group.}    
    \label{tab:groups_stats}
    \centering
    \setlength\tabcolsep{4pt}
    \begin{tabular}{clccl}
    \toprule
     \multirow{2}{*}{Id} & \multirow{2}{*}{Group name} & Sample & Unique & \multirow{2}{*}{Examples} \\
      &  & count & labels & \\
    \midrule
    1 & Inland water & 63 & 14 & reservoir, river, stream, canal, lake \\
    2 & Coastal water & 45 & 12 & coastal bay, ocean, bay, sea, coral reef \\
    3 & Coastal land & 39 & 16 & islet, coastal island, island, peninsula \\
    4 & Forest \& wetland & 30 & 9 & forest, mangrove forest, wetland, mangrove \\
    5 & Grass \& shrubland & 12 & 8 & scrubland, meadow, shrubland, hedge \\
    6 & Agriculture & 24 & 10 & orchard, farmyard, vineyard, agricultural field \\
    7 & Barren terrain & 40 & 15 & desert, cliff, glacier, sand desert, bare rock \\
    8 & Airport & 22 & 9 & airport apron, airport runway, military airfield \\
    9 & Transport & 31 & 15 & bridge, transit platform, parking lot, railway \\
    10 & Utilities & 32 & 17 & dam, pipeline, power plant, power substation \\
    11 & Industry & 34 & 10 & quarry, industrial facility, landfill, industrial area \\
    12 & Settlements & 32 & 13 & residential area, rural hamlet, rural village \\
    13 & Civic \& institutional & 51 & 25 & cemetery, school, university campus, prison \\
    14 & Recreation & 68 & 34 & sports pitch, urban park, golf course, marina \\
    \bottomrule
    \end{tabular}
    \vspace{-3mm}
\end{table}

\subsection{LLM prompting}

We selected the annotation model based on a qualitative experiment.
We ran a set of candidate multimodal LLMs on over 100 identical candidate masks and inspected their outputs, judging whether the assigned class label matched the polygon, whether the synonyms were usable as text queries, and whether the model rejected masks that were genuinely uninterpretable at 10\,m.
Mistral-Large-3 was the most consistent and we used it for the full corpus.

We prompt the model as a conservative remote-sensing annotator whose goal is to decide whether a candidate polygon provides useful supervision at Sentinel-2 resolution and, if so, to provide a class label and synonyms based on weak geospatial metadata. The prompt explicitly discourages hallucination, strongly penalizes ambiguity, and rejects administrative or legally defined concepts that are not visually grounded (e.g., legal borders).

Each prompt combines (i) a rendered Sentinel-2 RGB patch with the candidate polygon overlaid in a purple color, and (ii) structured metadata including geographic context (country/continent), a flag indicating whether the polygon covers the full image, the candidate \texttt{subtypes} tags from Overture or OpenStreetMap, and the polygon area as a percentage of the patch. The prompt instructs the model to treat \texttt{subtypes} as the primary signal for concept definition while using the image mainly to detect clear contradictions and to assess whether the mask is interpretable at the target resolution. Additionally, we add varing example outputs, counter examples, and specific instructions for the main failure cases from our visual review.

We constrain the model to generate \emph{JSON only} with a fixed output schema. Following a the chain-of-thought approach, the model first returns a short justification, followed by quality scores in $[0,1]$, and language annotations:
\begin{verbatim}
{
  "quality_reason": "...",
  "sample_quality_mask": 0.0,
  "class_label": "...",
  "class_synonyms": ["...", "..."]
}
\end{verbatim}
Every mask with a quality score larger than 0.5 is accepted and included in the dataset. We share the scores with the dataset, in case researchers want to use a more conservative threshold.
The scores are not continously distributed, so we treat them as a bernoulli trial instead.
The prompt further enforces output-length limits (e.g., short labels and short synonym phrases) and requires empty \texttt{class\_label} and \texttt{class\_synonyms} whenever the sample is rejected (e.g., both suitability scores are very low). 
\texttt{class\_synonyms} is constrained to a small set (2--4 items) describing the same physical concept in diverse noun phrases. The full prompt is shared in our code repository \url{https://github.com/IBM/TerraMesh-Masks}.

\subsection{Quality filtering}

We use several heuristics to filter out unsuited or incorrect source tags and increase the diversity of selected tags. Specifically, we sample up to 3 masks per image using the following sampling probability scores. If the LLM discards a sampled mask, we sample further candidates if available.

\textit{Frequency-based scoring.}
The first heuristic filter leverages previously validated reference masks to estimate spatially aware quality scores. We use the hierarchical, hexagonal geospatial indexing system H3 with resolution 3 ($\approx$110\,km in diameter) to group the annotation candidates. For a candidate mask from subtype $s$ in H3 cell $h$, we compute
\[
    p_{\mathrm{geo}}(s,h)=\frac{\mu \hat{x}_{\mathrm{global}}+n_{\mathrm{local}}\hat{x}_{\mathrm{local}}}{\mu+n_{\mathrm{local}}},
\]
where $\hat{x}_{\mathrm{global}}$ is the global score for subtype $s$, $\hat{x}_{\mathrm{local}}$ is the local score computed from reference masks within cell $h$, and $n_{\mathrm{local}}$ is the local observation count. The parameter $\mu$ controls the strength of the global prior and is set to 5 in our experiments.

We then combine the geographic prior and an area-based prior (using $\omega=0.8$) to compute the sampling score for candidate $s$ through
\[
    p_{\mathrm{freq}}(s,h)=\omega\,p_{\mathrm{geo}}(s,h)+(1-\omega)\,p_{\mathrm{area}}(s),
\]
where $p_{\mathrm{area}}(s)$ is an area-based score for subtype $s$. Candidates with $p_{\mathrm{geo}}<0.2$ are discarded: their score is set to $-\infty$, ensuring that they are never selected for downstream annotation. An exploration bonus is applied to rare subtypes so that underrepresented categories are not eliminated too early.

\textit{Spectral quality scoring.}
The second heuristic filter quantifies the spectral coherence of a candidate mask using the Normalized Difference Vegetation Index (NDVI), calculated using the S2 sample. The approach compares the NDVI statistics inside and outside the polygon to identify potentially incorrect annotations.
We distinguish three cases: (1)~For small masks covering less than 0.1\% of the image, the score is set to 1 by convention, because spectral discrimination is not meaningful at this small scale; (2)~Masks covering almost the entire patch with more than 85\% are scored by the normalized Shannon entropy of the NDVI histogram inside the mask, as inside-vs-outside contrast is uninformative; and (3)~for masks between 0.1\% and 85\% coverage, we measure spectral separability between inside and outside pixels using
    \[
        p_{\mathrm{spec}}=1 - \frac{abs(\mu_{\mathrm{in}}-\mu_{\mathrm{out}})}{\sqrt{\sigma_{\mathrm{in}}^2+\sigma_{\mathrm{out}}^2+\varepsilon}},
    \]
where $\mu_{\mathrm{in}}$, $\mu_{\mathrm{out}}$, $\sigma_{\mathrm{in}}$, and $\sigma_{\mathrm{out}}$ are the mean NDVI values and standard deviations inside and outside the mask, respectively. The assumption is that many correct annotations are uniform within the mask while having different scores to classes outside the mask.
To account for classes whose spectral signatures do not correctly indicate invalid masks (e.g., \textit{farmland} or \textit{airport}), a manually curated list of classes is downweighted before class selection.

\textit{Class-frequency downsampling.}
In addition to spatial filtering, the pipeline applies a global correction to downsample dominant subtypes like \textit{forest} or \textit{grassland}. Let $\pi_s$ denote the number of already selected masks of subtype $s$, and let $c$ be a fixed constant (here $c=0.02$). We define
\[
    p_{\mathrm{class}}=\min\!\left(1,\frac{c}{\max(\pi_s,\varepsilon)}\right),
\]
and use it to reweight the overall score. This regularizer does not force a perfectly uniform distribution, but it reduces sampling heavily overrepresented classes.

\textit{Final heuristic score.}
The final sampling score $p$ is computed as follows
\[
    p = p_{\mathrm{freq}} \cdot p_{\mathrm{spec}} \cdot p_{\mathrm{class}}.
\]
A hard gate is applied before the final multiplication: if $p_{\mathrm{geo}}<0.2$, the sample's probability is discarded. 
Conceptually, this score is designed to behave as a filter rather than a final label. The geographic prior removes subtype--region combinations that are consistently implausible, the spectral term suppresses masks that are likely misaligned with the actual image, and the class-frequency regularizer prevents single categories form dominating in the overall dataset. The result is a filtered candidate pool that is much smaller, substantially cleaner, and more diverse to support open-vocabulary segmentation at scale.

\section{Training details}
\label{sec:training_details}

\textit{Multimodal and (S2) models.}
The two model families differ only in which input modalities are available during pretraining, not in architecture, schedule, or training samples.
The multimodal models receive Sentinel-2 L2A, Sentinel-2 RGB, Sentinel-1 GRD, Sentinel-1 RTC, and DEM as input domains.
The (S2) models receive only Sentinel-2 L2A and Sentinel-2 RGB.
During training, we sample 392 input tokens for the multimodel models, which represents two full input modalities, and 196 tokens for the (S2) models.
Both versions predict the same target domain of binary-mask tokens and use the same text query as additional input tokens. We do not sample the target tokens and always use all 196 tokens.

\textit{NDVI-based filtering during pretraining.}
Beyond the filtering applied when constructing the corpus, we apply a second filter in the pretraining pipeline.
We keep only masks whose NDVI agreement score is at least 0.4, which removes roughly 23\% of the released training masks.
This threshold is deliberately stricter than what we use for the released dataset, where we keep every mask above the LLM quality score of 0.5 and publish the agreement scores so that others can choose their own operating point.
Filtering at pretraining time rather than at release keeps the dataset broad while giving the model a cleaner training signal.

\textit{Class-label reweighting.}
The distribution over class labels in TerraMesh-Masks is long tailed, since common concepts such as \textit{forest} or \textit{residential area} appear far more often than \textit{funicular} or \textit{coral reef}.
We counteract this with stochastic repetition at the sample level.
Each class label $c$ is assigned a repeat factor $\lambda_c$, which represents the expected repetition per epoch of a sample with this label.
Factors range from 0.1 to a cap of 500, so samples of the most frequent labels are often dropped while the rarest are repeated heavily.
This flattens the label distribution in both directions, which matters for open-vocabulary segmentation where the tail of the label distribution is what the model is expected to generalize over.

\section{Additional results}
\label{sup:add_results}

\begin{figure}
\begin{center}
        \includegraphics[width=0.24\linewidth]{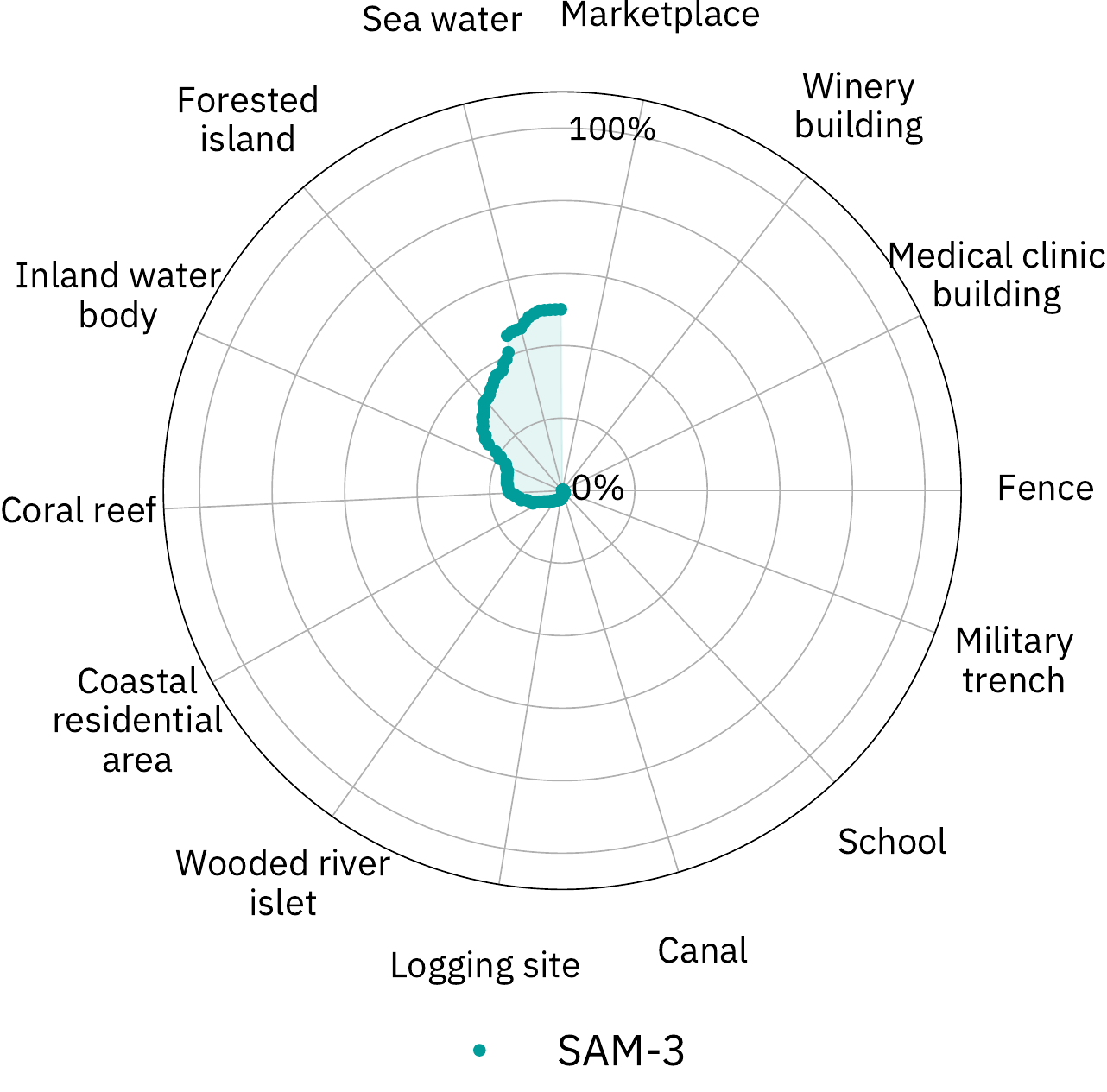}
    \hfill
    \includegraphics[width=0.24\linewidth]{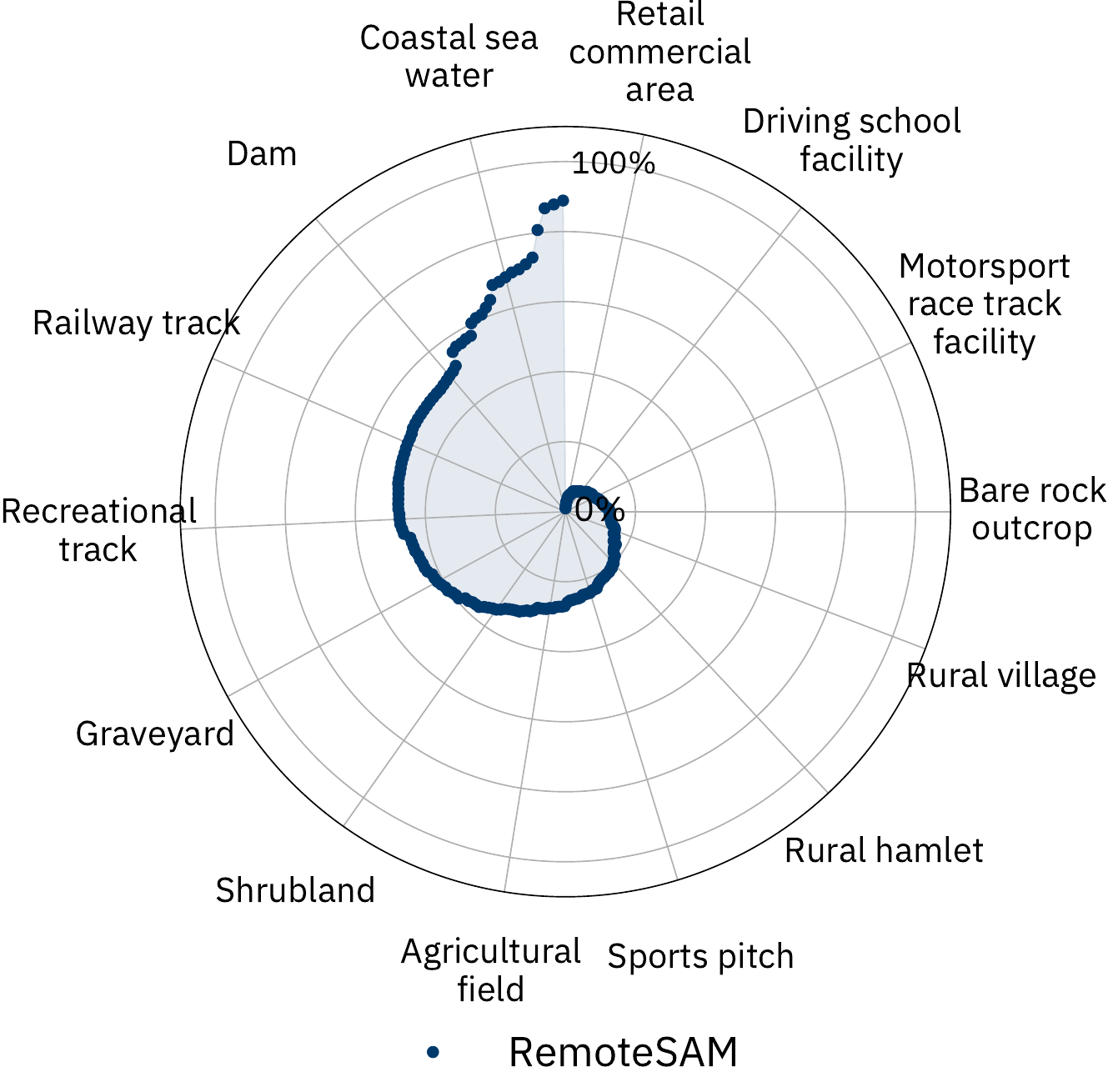}
    \hfill
    \includegraphics[width=0.24\linewidth]{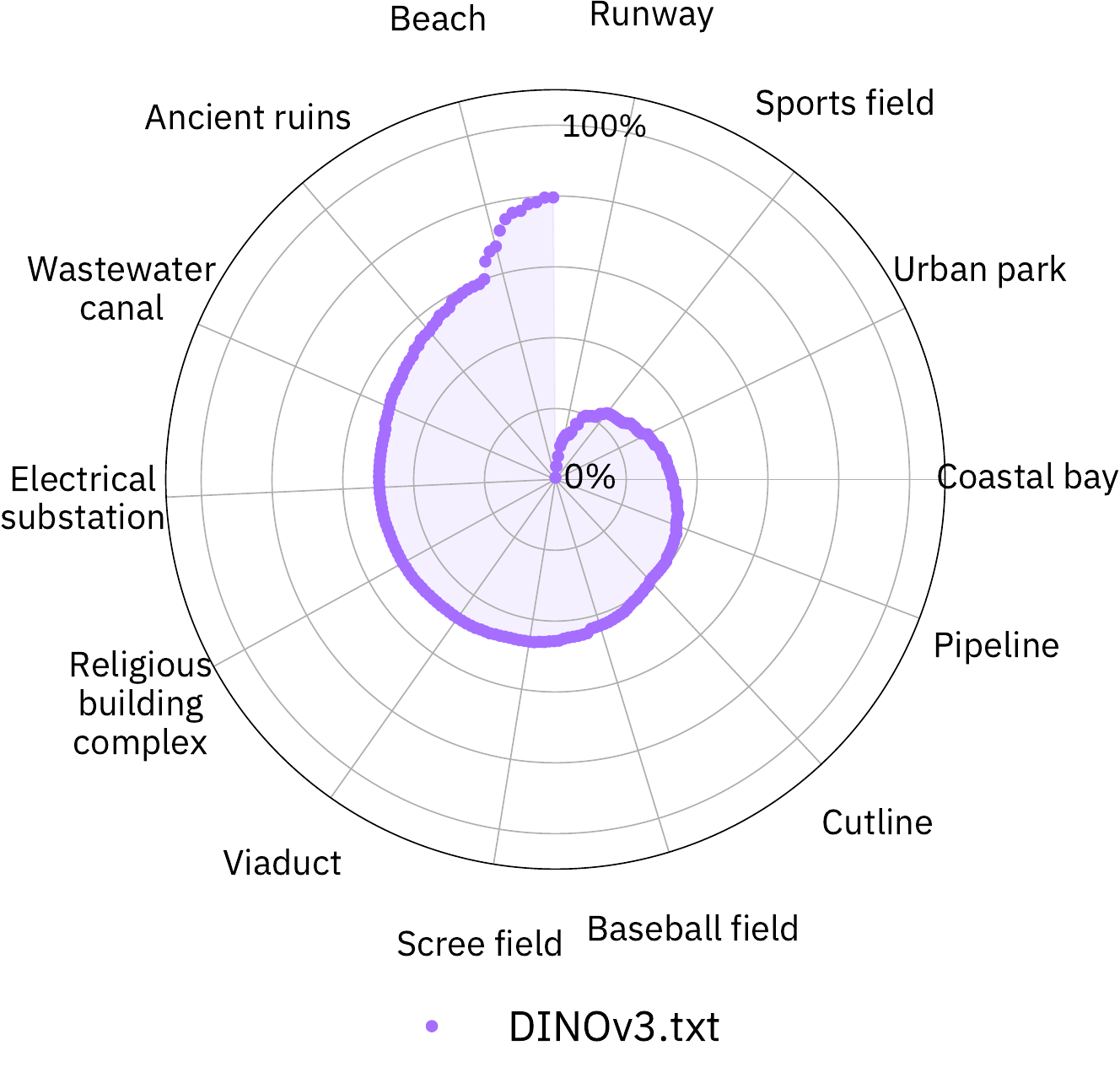}
    \hfill
    \includegraphics[width=0.24\linewidth]{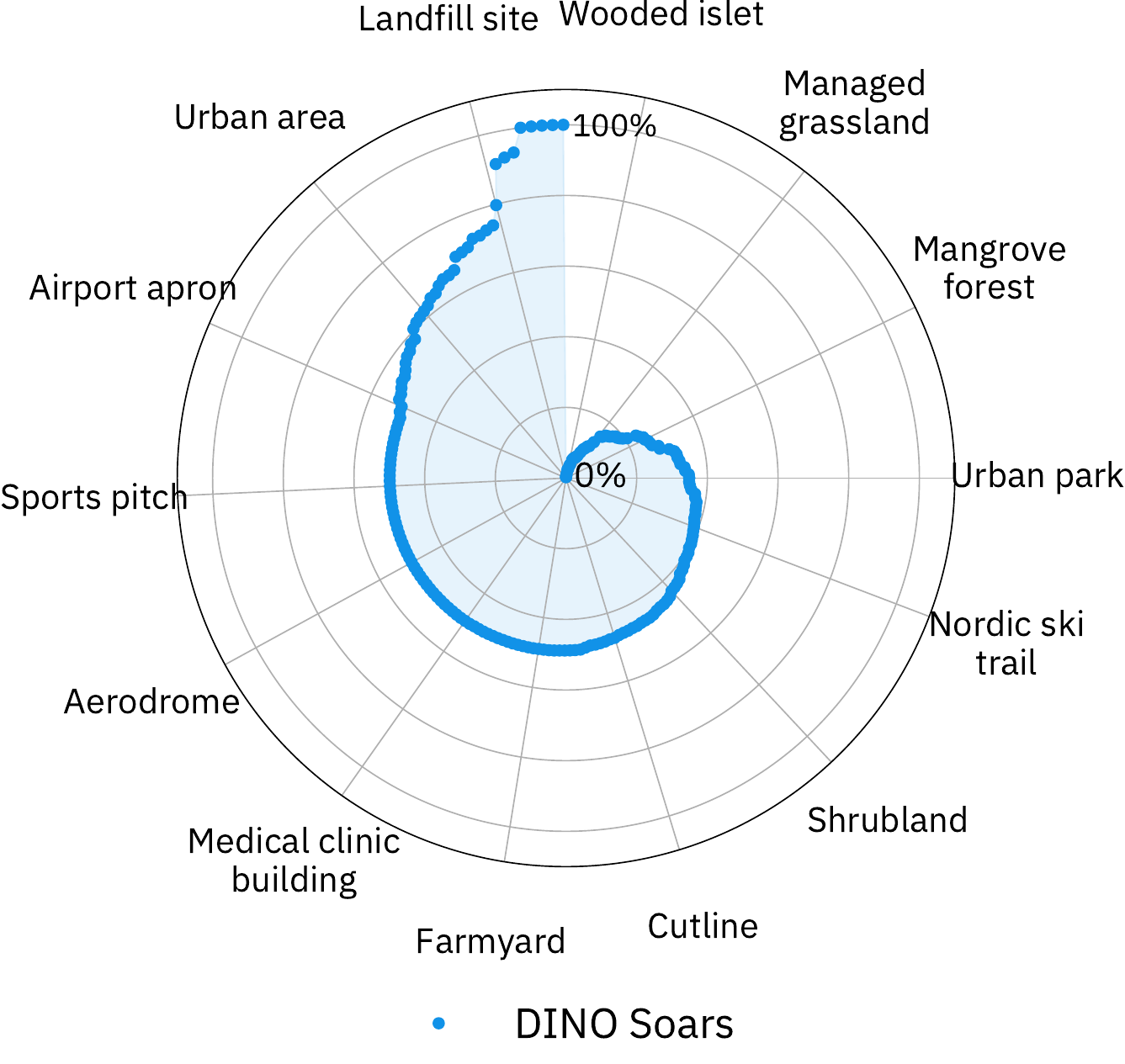}

    \includegraphics[width=0.24\linewidth]{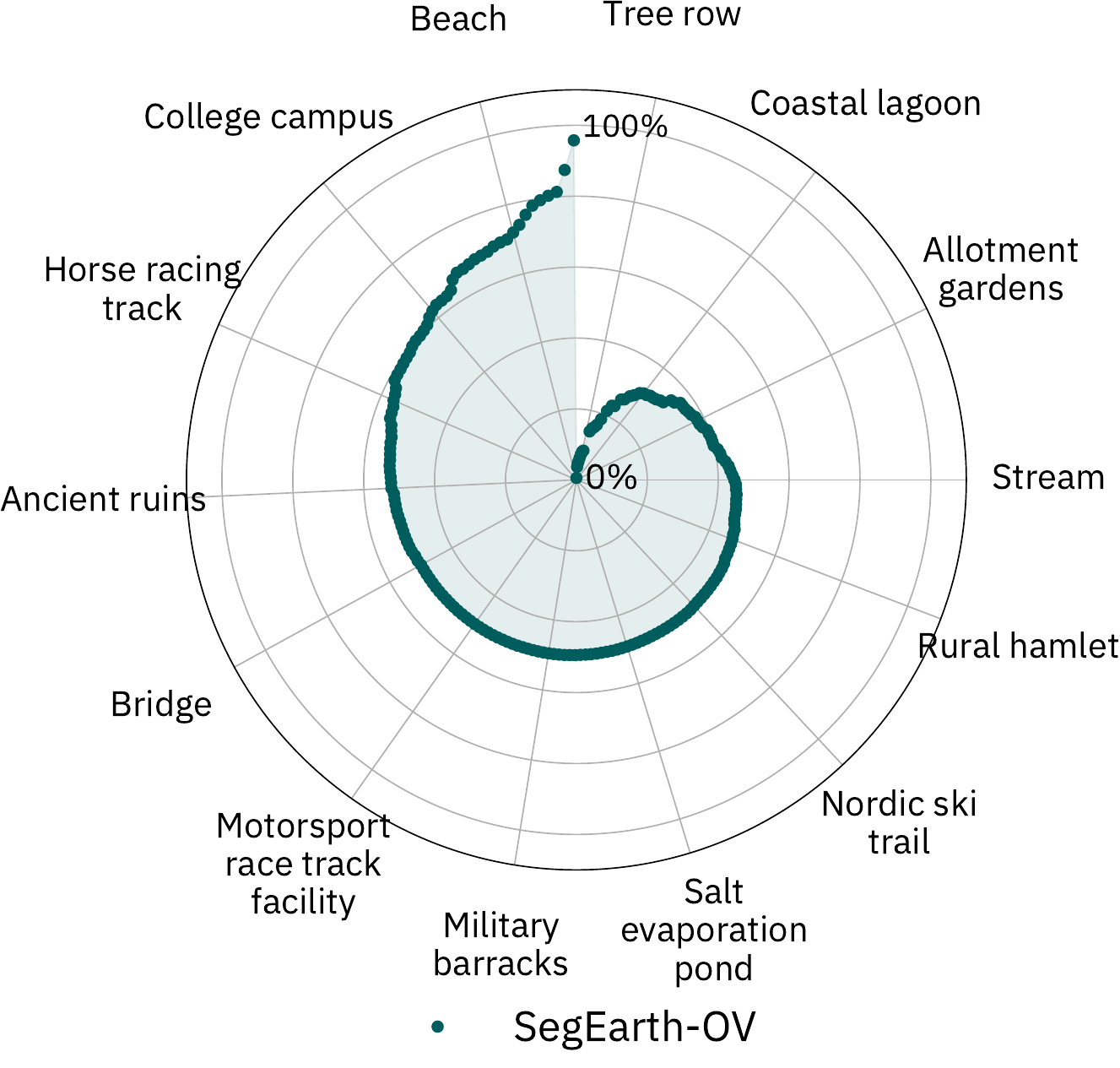}
    \includegraphics[width=0.24\linewidth]{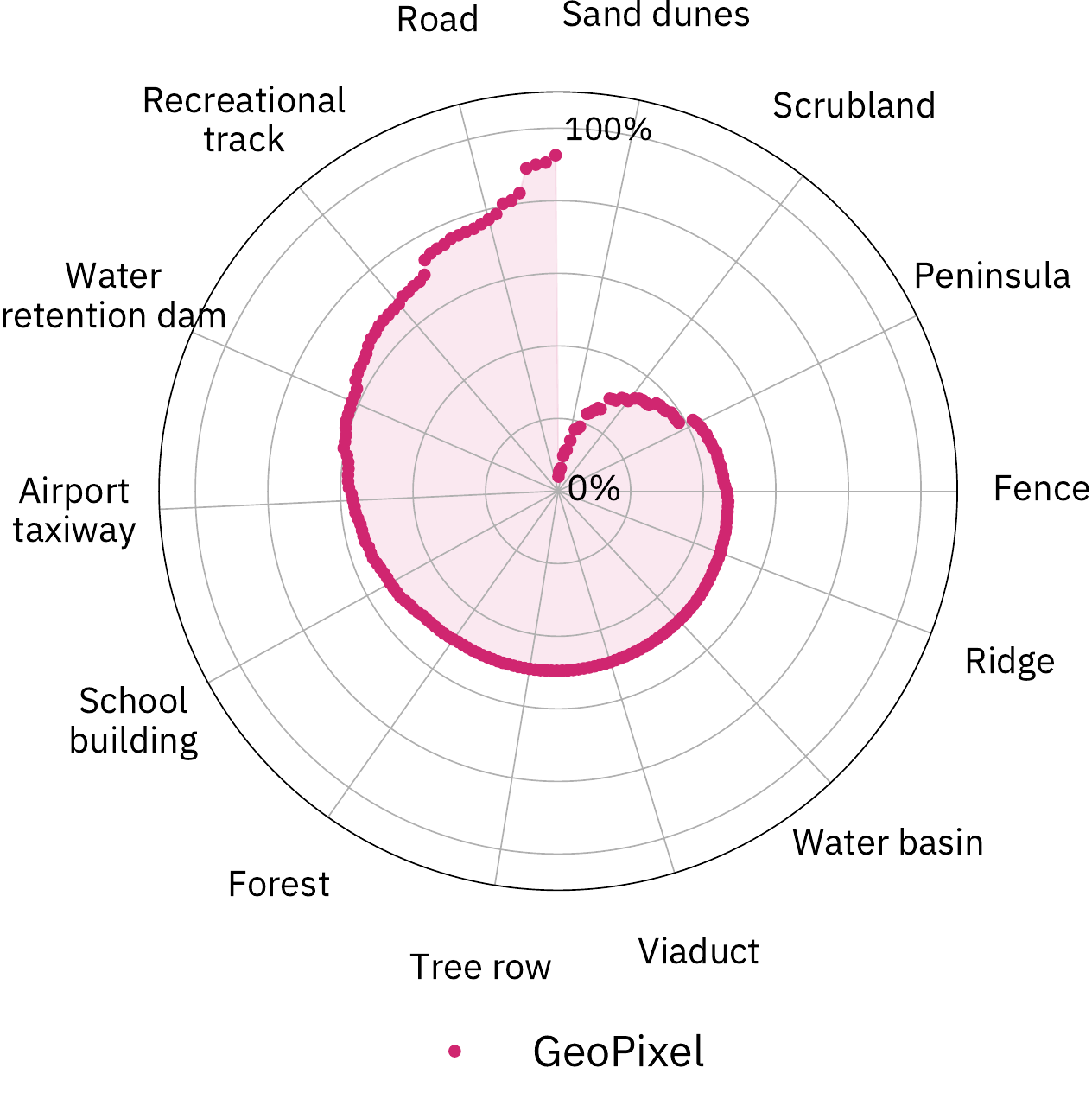}
    \includegraphics[width=0.24\linewidth]{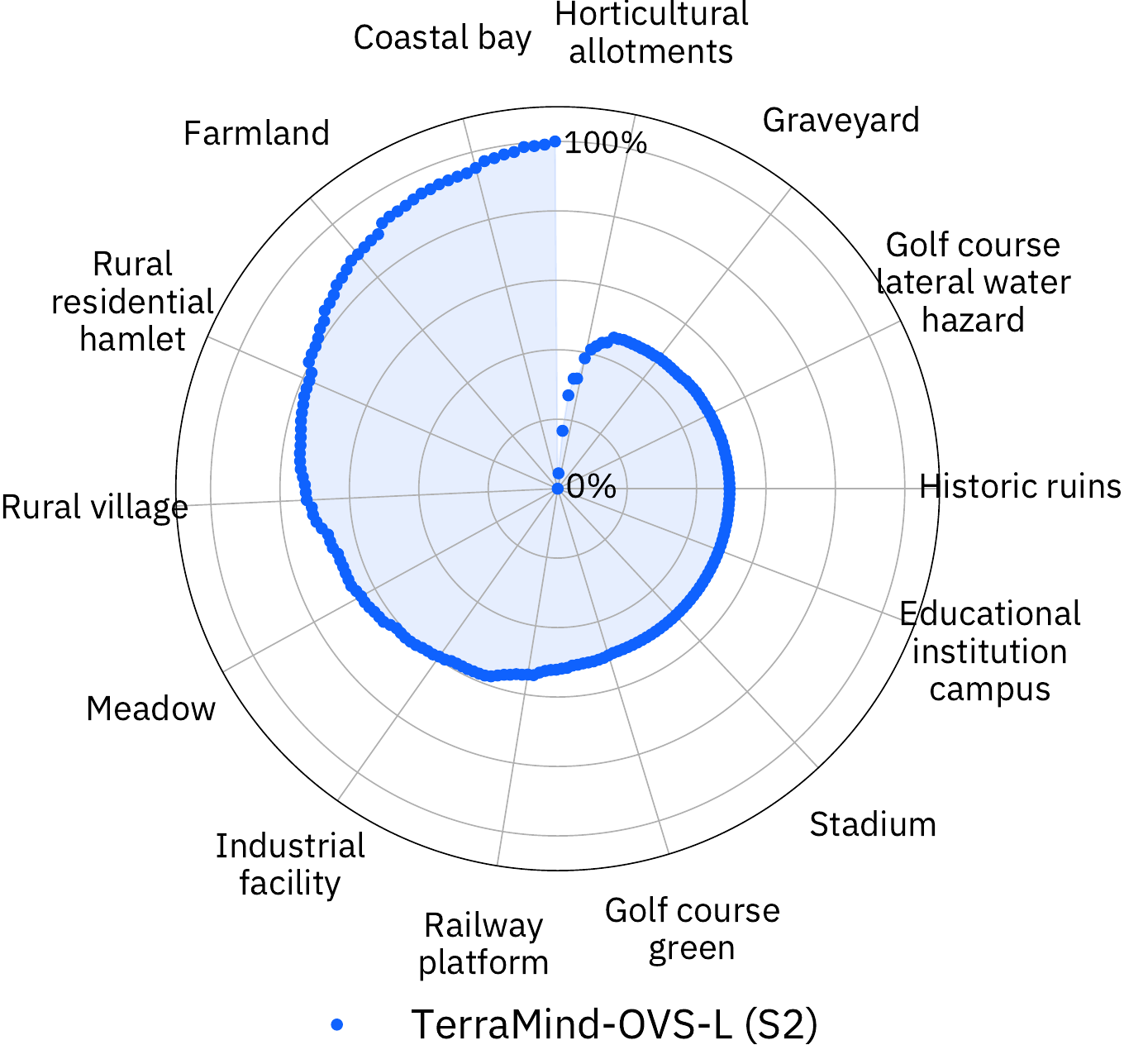}

    \caption{Label-wise mIoU results of all 207 labels in the eval set, sorted per model.}
    \label{fig:class_miou}
\end{center}
\end{figure}

\textit{Class-wise results.}
Beyond reporting grouped results, we provide per-class mIoU scores in Fig.~\ref{fig:class_miou}. The sample performances are grouped by their class labels and then averaged. However, due to the diversity of labels, many scores are based on a single sample. Similar to the aggregated scores, SAM-3 shows the worst performance due to many missing predictions. Many false positive predictions from SAM-3 and DINOv3.txt contribute to a generally lower mIoU. GeoPixel performs quite well on smaller objects which is highlighted by the label examples. These classes are well covered in the GeoPixel training dataset~\cite{shabbir2025geopixel}. In contrast, TerraMind-OVS-L performs much better on land-cover-related classes.

\begin{figure}
    \centering
    \includegraphics[width=\linewidth]{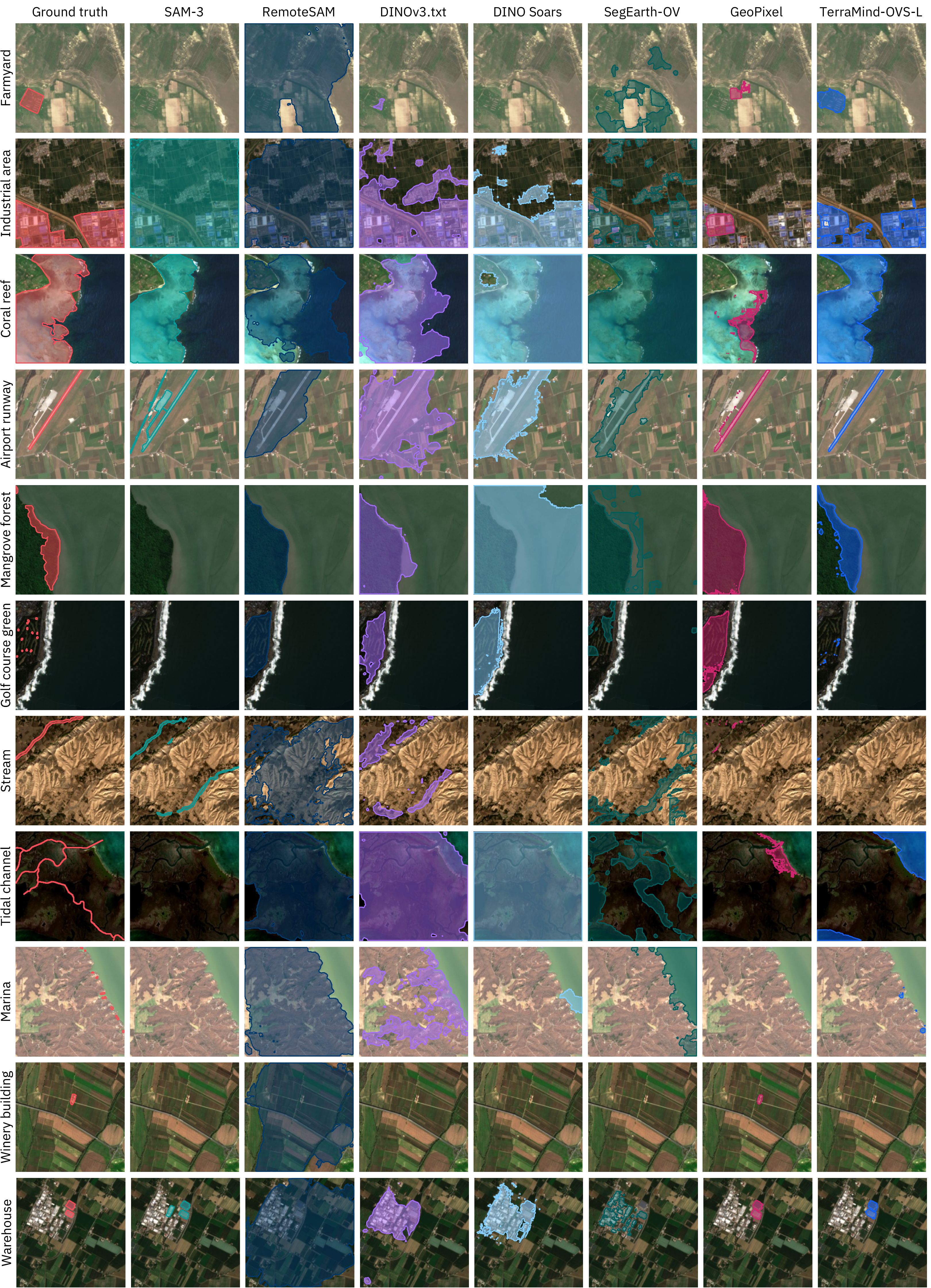}
    \caption{Qualitative comparison on examples from TerraMesh-Masks-Eval. SAM- and DINOv3-based models often lead towards overprediction or no prediction. TerraMind-OVS performs well on larger land-cover related labels, while GeoPixel is much better in spotting objects.}
    \label{fig:supp-predictions}
\end{figure}

\textit{Qualitative results.}
We provide additional predictions in Fig.~\ref{fig:supp-predictions}. The samples are selected representatively to highlight different strengths and weaknesses of the models. For example, TerraMind-OVS is very good at detecting land airport related tasks, while failing on small rivers and streams.

\subsection{Per-category performance}
Using a grouping of the label space into 14 coarse semantic groups, we observe a clear structure in model behavior (Tab.~\ref{tab:supp-group-results}). Across model sizes, broad land-cover and vegetation classes are generally segmented more reliably than thin, fragmented, or highly structured targets. This pattern is consistent with the visual character of the task and with the spatial resolution of Sentinel-2: extended regions with coherent texture are substantially easier than narrow infrastructure elements or boundaries with limited visual support.

Table~\ref{tab:supp-group-results} includes a per-group comparison of TerraMind-OVS-L using RGB and Sentinel-2 inputs. We see large differences between the multimodal and S2 version depending on the group. Within each version, the multispectral inputs slightly improve the performance in most groups, but only within the confidence intervals. 
In contrast, other results from literature showed significant improvements after integrating multispectral inputs into a VLM~\cite{marimo2025beyond}. This is likely due to the nature of many man-made objects and land-cover classes that are clearly visible in RGB.

\begin{table}[]
    \caption{IoU results on TerraMesh-Masks-Eval per group. All models are using RGB inputs, except for the last two rows in each block, which show TerraMind-OVS-L with S2 L2A inputs.}
    \label{tab:supp-group-results}
    \centering
    \scriptsize
    \setlength\tabcolsep{2.5pt}
    \begin{tabularx}{\textwidth}{p{2mm} X *{15}{r}}
    \toprule
    & Model & \rotate{Inland water} & \rotate{Coastal water} & \rotate{Coastal land} & \rotate{Forest \& wet.} & \rotate{Grass \& shrub.} & \rotate{Agriculture} & \rotate{Barren terrain} & \rotate{Airport} & \rotate{Transport} & \rotate{Utilities} & \rotate{Industry} & \rotate{Settlements} & \rotate{Civic \& inst.} & \rotate{Recreation} & Mean \\
    \midrule
    \multirow{15}{*}{\rotate{IoU$_{pos}$}}
    & SAM-3 & 20.49 & 43.58 & 22.77 & 3.48 & 5.84 & 5.26 & 12.91 & 4.66 & 0.00 & 0.14 & 10.90 & 11.97 & 1.08 & 0.77 & 10.28 \\
    & RemoteSAM & 6.24 & 19.98 & 27.29 & 32.51 & \textbf{28.95} & 17.89 & 36.96 & 13.92 & 5.04 & 4.51 & 9.71 & 13.76 & 2.48 & \underline{9.50} & 16.34 \\
    & DINOv3.txt & 11.22 & 42.05 & 23.41 & 26.12 & 8.32 & 17.03 & 21.80 & 12.58 & 4.57 & 5.09 & 19.76 & 26.55 & 7.90 & 4.96 & 16.53 \\
    & DINO Soars & 11.79 & 58.89 & 21.89 & 39.29 & 16.10 & 13.99 & 42.61 & 16.97 & 1.52 & 5.90 & 20.56 & 26.47 & 3.54 & 2.96 & 20.18 \\
    & SegEarth-OV & 16.82 & 57.85 & 23.32 & 39.14 & 18.62 & 18.89 & 46.84 & 10.99 & 3.89 & 5.76 & 18.36 & 33.13 & 7.20 & 5.87 & 21.91 \\
    & GeoPixel & 9.60 & 19.65 & 19.51 & 20.01 & 7.26 & 16.59 & 13.76 & 19.13 & \textbf{16.56} & 4.47 & 25.32 & 20.26 & \underline{12.46} & \textbf{11.26} & 15.42 \\
    & TerraMind-OVS-S & 24.63 & \underline{83.82} & 50.13 & 47.12 & 13.65 & 25.20 & 44.18 & 16.02 & 2.46 & 3.09 & 13.87 & 41.02 & 1.71 & 3.61 & 26.46 \\
    & TerraMind-OVS-S (S2) & 27.07 & \textbf{84.62} & 49.39 & 45.67 & 19.53 & 24.07 & 42.99 & 16.40 & 3.47 & 3.13 & 19.66 & 42.84 & 1.26 & 4.43 & 27.47 \\
    & TerraMind-OVS-B & 29.52 & 82.90 & 59.67 & \textbf{53.67} & 21.09 & 35.18 & \underline{52.76} & 24.97 & 3.53 & 5.38 & 26.19 & 58.04 & 3.32 & 5.88 & 33.01 \\
    & TerraMind-OVS-B (S2) & 31.87 & 83.02 & 58.31 & 50.40 & 21.85 & 35.02 & 43.75 & 23.17 & 6.73 & 8.06 & 26.67 & 56.77 & 8.06 & 4.96 & 32.76 \\
    & TerraMind-OVS-L & 31.85 & 74.20 & \textbf{64.08} & 50.77 & 18.65 & \underline{41.87} & 44.23 & \underline{34.95} & 8.34 & \underline{14.91} & \underline{45.80} & 50.96 & 11.51 & 8.71 & 35.77 \\
    & TerraMind-OVS-L (S2) & 29.55 & 82.56 & 62.96 & 50.96 & 24.08 & 34.80 & 52.56 & 29.00 & \underline{11.27} & 11.85 & 33.40 & \underline{61.66} & 10.11 & 7.46 & 35.87 \\
    \cmidrule{2-17}
    & TM-OVS-L, S2 input & \underline{32.01} & 74.07 & \underline{63.84} & 50.91 & 20.23 & \textbf{42.57} & 45.30 & \textbf{35.26} & 8.39 & \textbf{14.93} & \textbf{45.89} & 51.59 & 11.79 & 8.61 & \textbf{36.10} \\
    & TM-OVS-L (S2), S2 inp. & \textbf{32.07} & 82.66 & 63.34 & \underline{52.06} & \underline{24.69} & 33.97 & \textbf{53.02} & 29.51 & 9.67 & 7.83 & 31.11 & \textbf{64.53} & \textbf{13.04} & 6.02 & \underline{35.96} \\
    \midrule
    \multirow{15}{*}{\rotate{mIoU}}
    & SAM-3 & 11.92 & 35.27 & 15.02 & 17.17 & 19.08 & 8.63 & 24.00 & 7.91 & 0.46 & 0.97 & 7.29 & 8.40 & 1.46 & 1.37 & 11.35 \\
    & RemoteSAM & 29.98 & 22.02 & 42.03 & 39.45 & 30.71 & 24.91 & 31.38 & 32.37 & 34.30 & 38.42 & 26.30 & 17.80 & 18.29 & 37.99 & 30.42 \\
    & DINOv3.txt & 39.44 & 35.64 & 31.04 & 45.05 & 31.35 & 45.24 & 30.60 & 30.47 & 45.97 & 43.96 & 51.03 & 51.76 & 47.44 & 44.17 & 40.94 \\
    & DINO Soars & 43.05 & 46.63 & 20.69 & 45.95 & 22.74 & 33.35 & 52.07 & 47.04 & 47.56 & 50.36 & 54.68 & 44.77 & 46.94 & 47.10 & 43.07 \\
    & SegEarth-OV & 49.92 & 53.58 & 36.78 & 44.40 & 30.79 & 42.48 & 39.19 & 49.32 & 49.65 & 51.13 & 52.96 & 59.39 & 52.26 & 49.92 & 47.27 \\
    & GeoPixel & 50.54 & 34.40 & 50.35 & 44.85 & 33.73 & 50.59 & 32.04 & 53.93 & \textbf{57.67} & 50.68 & 59.71 & 54.04 & \underline{55.14} & \textbf{53.65} & 48.67 \\
    & TerraMind-OVS-S & 60.79 & \underline{88.89} & 69.38 & 64.02 & 40.17 & 57.72 & 52.11 & 53.52 & 50.68 & 50.25 & 53.96 & 66.44 & 49.90 & 50.89 & 57.76 \\
    & TerraMind-OVS-S (S2) & 62.10 & \textbf{89.40} & 68.64 & 63.09 & 41.61 & 56.71 & 52.30 & 51.15 & 51.08 & 50.50 & 57.23 & 67.44 & 49.67 & 51.16 & 58.01 \\
    & TerraMind-OVS-B & 63.50 & 87.87 & 73.36 & \textbf{69.00} & \textbf{44.30} & 62.92 & \textbf{58.84} & 58.92 & 51.26 & 51.68 & 60.49 & 76.41 & 50.39 & 51.98 & 61.50 \\
    & TerraMind-OVS-B (S2) & 64.69 & 86.86 & 73.51 & 66.67 & 42.71 & 62.59 & 52.79 & 56.68 & 52.78 & 53.01 & 60.71 & 75.69 & 52.97 & 51.58 & 60.95 \\
    & TerraMind-OVS-L & 64.65 & 79.14 & \textbf{78.67} & 67.74 & 41.77 & \underline{67.35} & 51.17 & \underline{64.15} & 53.64 & \textbf{56.39} & \underline{71.72} & 71.36 & 54.28 & \underline{53.34} & 62.53 \\
    & TerraMind-OVS-L (S2) & 63.53 & 87.82 & 78.18 & 66.62 & 42.74 & 63.15 & 55.44 & 60.73 & \underline{55.13} & 54.90 & 64.63 & \underline{78.21} & 53.94 & 52.80 & \underline{62.70} \\
    \cmidrule{2-17}
    & TM-OVS-L, S2 input & \underline{64.74} & 79.02 & \underline{78.44} & \underline{67.82} & 41.54 & \textbf{67.69} & 51.71 & \textbf{64.30} & 53.65 & \underline{56.38} & \textbf{71.77} & 71.70 & 54.33 & 53.19 & 62.59 \\
    & TM-OVS-L (S2), S2 inp. & \textbf{64.89} & 87.82 & 78.31 & 67.63 & \underline{43.46} & 62.53 & \underline{57.06} & 61.55 & 54.33 & 52.87 & 63.25 & \textbf{79.81} & \textbf{55.52} & 51.99 & \textbf{62.93} \\
    \bottomrule
    \end{tabularx}
\end{table}

\subsection{Effect of mask size}

\begin{figure}[]
    \centering
    \includegraphics[width=0.8\linewidth]{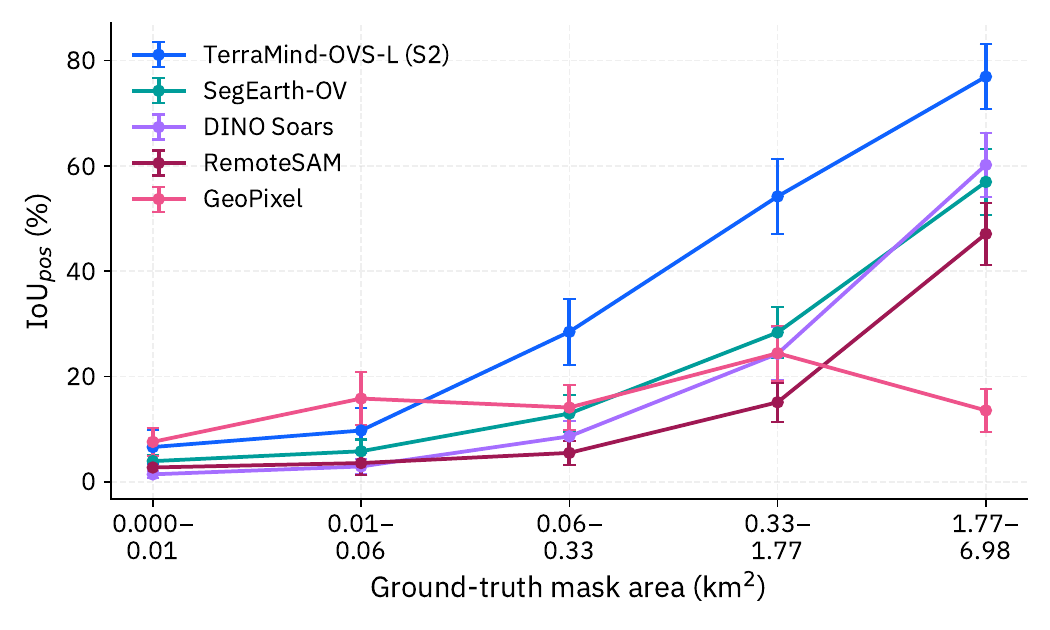}
    \caption{IoU$_{pos}$ against ground-truth mask area on TerraMesh-Masks-Eval for the EO-specific models, binned into log-spaced quintiles. Error bars show 95\% confidence intervals.}
    \label{fig:area_vs_performance}
\end{figure}

Since our label space mixes extended land-cover regions with small individual objects, we relate performance to the size of the target mask in Fig.~\ref{fig:area_vs_performance}.
We bin the evaluation samples into log-spaced quintiles of ground-truth mask area, roughly 105 samples per bin, and report IoU$_{pos}$ within each bin. We restrict the comparison to the EO-specific baselines.
Mask area is the dominant factor for all methods.
TerraMind-OVS-L (S2) improves from 6.6\% IoU$_{pos}$ on the smallest bin, below 0.01\,km$^2$, to 76.9\% on the largest ones.
SegEarth-OV, DINO Soars, and RemoteSAM follow the same monotone trend at a lower level.
GeoPixel behaves differently and its best results are in the middle bins.
It is the strongest model on the two smallest bins, at 7.6\% and 15.8\% IoU$_{pos}$, consistent with its training data of small annotated objects in high-resolution imagery. It is the only model that degrades on the largest targets of land cover classes, falling to 13.5\%.
This confirms that the remaining headroom on our benchmark is concentrated on small and thin structures, which are the hardest to resolve at 10\,m.

\subsection{Effect of strong distribution shifts}

\begin{figure}[h]
    \centering
    \includegraphics[width=0.8\linewidth]{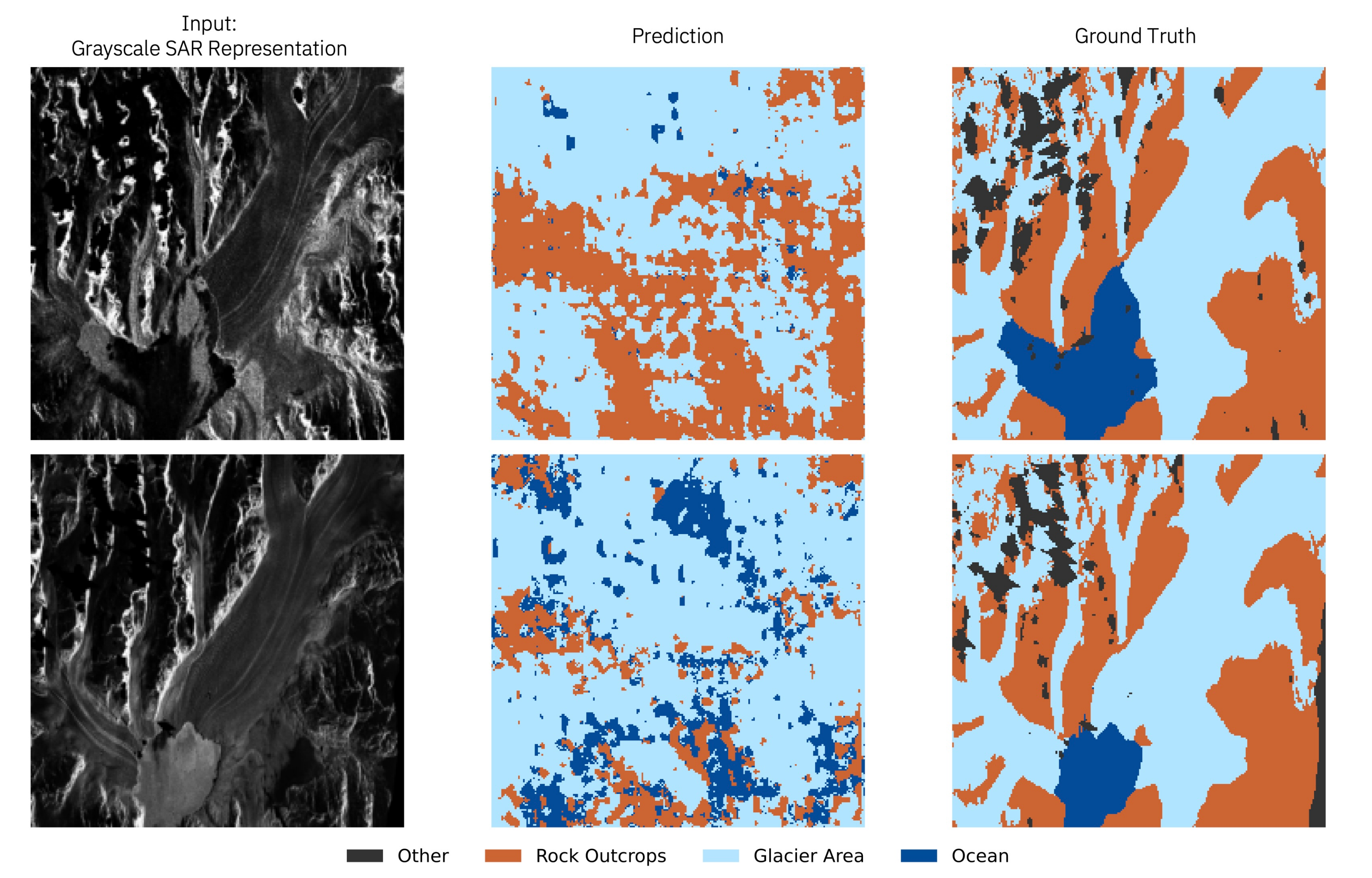}
    \caption{Predictions on CaFFe with TerraMind-OVS-S.}
    \label{fig:caffe}
\end{figure}

With the evaluation of multi-label datasets, we also investigate the effect of strong distribution shifts on model performance using the example of the CaFFe dataset. CaFFe provides data on glacier calving, derived from diverse SAR images represented as single channel grayscale images. We therefore expect no benefit in using SAR-specific input channels of TerraMind-OVS. Instead, for all models across the benchmark, we stack three identical grayscale images and feed this as RGB representation to the models. All tested models struggle to compensate for this strong distribution shift. In Fig.~\ref{fig:caffe}, we provide examples of the predictions of TerraMind-OVS-S on CaFFe test data. We observe significant artefacts and problems in identifying segments, which are further amplified with most other benchmarked models. 

Overall, we observe strong generalizability across diverse concepts, such as those in TerraMesh-Masks-Eval. However, we find that this generalizability is limited by very strong distribution shifts that the model is not suitably adapted to. 

\end{document}